%% file: main.tex
\documentclass{article}

\usepackage[margin=1.1in]{geometry} %
\usepackage[font={small, sf}]{caption}
\usepackage[hang,flushmargin]{footmisc}

\usepackage[utf8]{inputenc} 
\usepackage[T1]{fontenc}    
\usepackage{hyperref}       
\usepackage{url}            
\usepackage{booktabs}       
\usepackage{amsfonts}       
\usepackage{nicefrac}       
\usepackage{microtype}      
\usepackage{optidef}
\usepackage{float}
\usepackage{placeins}
\usepackage{empheq}
\usepackage{amsthm}
\usepackage{aliascnt}
\usepackage{xcolor}
\usepackage{caption}
\usepackage[ruled]{algorithm2e}
\usepackage[export]{adjustbox}
\usepackage[labelfont=bf,skip=0pt, singlelinecheck=false]{caption}

\newcommand\blfootnote[1]{%
  \begingroup
  \renewcommand\thefootnote{}\footnote{#1}%
  \addtocounter{footnote}{-1}%
  \endgroup
}
\newaliascnt{eqfloat}{equation}
\newfloat{eqfloat}{h}{eqflts}
\floatname{eqfloat}{Equation}
\newcommand*{\ORGeqfloat}{}
\let\ORGeqfloat\eqfloat
\def\eqfloat{%
  \let\ORIGINALcaption\caption
  \def\caption{%
    \addtocounter{equation}{-1}%
    \ORIGINALcaption
  }%
  \ORGeqfloat
}

\usepackage{thmtools, thm-restate}

\usepackage{authblk}
\title{The Ramanujan Machine: Automatically Generated Conjectures on Fundamental Constants}
\author[$\dag$1]{Gal Raayoni}
\author[$\dag$1]{Shahar Gottlieb}
\author[1]{George Pisha}
\author[1]{Yoav Harris}
\author[1]{Yahel Manor}
\author[2]{Uri Mendlovic}
\author[1]{Doron Haviv}
\author[1]{Yaron Hadad}
\author[1]{Ido Kaminer}
\affil[1]{Technion - Israel Institute of Technology, Haifa 3200003, Israel}
\affil[2]{Google Inc., Tel Aviv 6789141, Israel}
\affil[$\dag$]{Equal contribution}
\begin{document}

\maketitle

\begin{abstract}
Fundamental mathematical constants like $e$ and $\pi$ are ubiquitous in diverse fields of science, from abstract mathematics and geometry to physics, biology and chemistry. Nevertheless, for centuries new mathematical formulas relating fundamental constants have been scarce and usually discovered sporadically. In this paper we propose a novel and systematic approach that leverages algorithms for deriving mathematical formulas for fundamental constants and help reveal their underlying structure. Our algorithms find dozens of well-known as well as previously unknown continued fraction representations of $\pi$, $e$, Catalan's constant, and values of the Riemann zeta function. Two example conjectures found by our algorithm and in retrospect simple to prove are:
\begin{equation*}
     \frac{e}{e-2} = 4 - \frac{1}{5-\frac{2}{6-\frac{3}{7-\frac{4}{8-...}}}}
    \quad\quad,\quad\quad
    \frac{4}{3\pi - 8} = 3-\frac{1\cdot1}{6-\frac{2\cdot3}{9-\frac{3\cdot5}{12-\frac{4\cdot7}{15-...}}}}
\end{equation*}
Two example conjectures found by our algorithm and so far unproven are:
\begin{equation*}
     \frac{24}{\pi^2} = 2 + 7\cdot 0\cdot 1+ \frac{8\cdot1^4}{2 + 7\cdot 1\cdot 2 + \frac{8\cdot2^4}{2 + 7\cdot 2\cdot 3 + \frac{8\cdot3^4}{2 + 7\cdot 3\cdot 4 + \frac{8\cdot4^4}{..}}}}
    \quad\quad,\quad\quad
    \frac{8}{7 \zeta\left(3\right)} = 1\cdot 1 - \frac{1^6}{3\cdot 7 - \frac{2^6}{5\cdot 19 - \frac{3^6}{7\cdot 37 - \frac{4^6}{..}}}}
\end{equation*}
We present two algorithms that proved useful in finding conjectures: a variant of the Meet-In-The-Middle (MITM) algorithm and a Gradient Descent (GD) tailored to the recurrent structure of continued fractions. Both algorithms are based on matching numerical values and thus they conjecture formulas without providing proofs and without requiring any prior knowledge on any underlying mathematical structure. This approach is especially attractive for fundamental constants for which no mathematical structure is known, as it reverses the conventional approach of sequential logic in formal proofs. Instead, our work supports a different conceptual approach for research: computer algorithms utilizing numerical data to unveil mathematical structures, thus trying to play the role of intuition of great mathematicians of the past, providing leads to new mathematical research.

\end{abstract}

\blfootnote{Code available at: \url{http://www.ramanujanmachine.com/} and the git links inside}

\newpage

\section{Introduction} \label{sec:Intro}

Fundamental mathematical constants such as $e$, $\pi$, the golden ratio $\varphi$, and many others play an instrumental part in diverse fields such as geometry, number theory, calculus, fundamental physics, biology, and ecology \cite{finch2004reviews}. Throughout history simple formulas of fundamental constants symbolized simplicity, aesthetics, and mathematical beauty \cite{questforpi, sloane2003line, integerencyclopedia}. A couple of well-known examples include Euler’s identity $e^{i \pi}+1=0$ and the continued fraction representation of the Golden ratio:
\begin{equation} \label{GoldenRatio}
\varphi=1 + \frac{1}{1+\frac{1}{1+\frac{1}{1+\dots}}}.
\end{equation}

The discovery of such Regular Formulas (RFs)\footnote{By regular formulas we refer to any mathematical expression that can be encapsulated using a computable expression \cite{turing1936computable}, even if it may seem infinite in nature} is often sporadic and considered an act of mathematical ingenuity or profound intuition. One prominent example is Gauss' ability to see meaningful patterns in numerical data that led to new fields of analysis such as elliptic and modular functions and led to the hypothesis of the Prime Number Theorem. He is even famous for saying: ``I have the result, but I do not yet know how to get it'' \cite{borwein2008mathematics}, which emphasizes the role of identifying patterns and RFs in data as enabling acts of mathematical discovery.

In a different field but a similar manner, Johannes Rydberg's discovery of his formula of hydrogen spectral lines \cite{bohr1954rydberg}, resulted from his data analysis of the spectral emission by chemical elements: $\lambda^{-1}=R_{H}(n^{-2}_1 - n^{-2}_2)$, where $\lambda$ is the emission wavelength, $R_{H}$ is the Rydberg constant, $n_{1}$ and $n_{2}$ are the upper and lower quantum energy levels, respectively. This insight, emerging directly from identifying patterns in the data, had profound implications on modern physics and quantum mechanics.

Unlike measurements in physics and all other sciences, most \textit{mathematical} constants can be calculated to an arbitrary precision\footnote{Exceptions for this are constructions such as the Chaitin's constant \cite{raatikainen1998interpreting}} (number of digits) with an appropriate formula, thus providing an \textit{absolute ground truth}. In this sense, mathematical constants contain an unlimited amount of data (e.g., the infinite sequence of digits in an irrational number), which we use as ground truth for finding new RFs. Since the fundamental constants are universal and ubiquitous in their applications, finding such patterns can reveal new mathematical structures with broad implications, e.g., the Rogers-Ramanujan continued fraction (which has implications on modular forms) and the Dedekind $\eta$ and j functions \cite{shimura1973modular, WolframContFrac}. Consequently, having \textit{systematic} methods to derive new RFs can help research in many fields of science. 

In this paper, we establish a novel method to learn mathematical relations between constants and present a list of conjectures found using this method. While the method can be leveraged for many forms of RFs, we demonstrate its potential with equations involving polynomial continued fractions (PCFs) \cite{cuyt2008handbook} in which the partial numerators and denominators follow closed-form polynomials:
\begin{equation} \label{PCFExample}
    x = a_{0} + \frac{b_{1}}{a_{1} + \frac{b_{2}}{a_{2} + \frac{b_{3}}{a_{3} + \dots}}},
\end{equation}
where $a_{n}, b_{n} \in \mathbb{Z}$ for $n=1,2,\dots$ are partial numerators and denominators, respectively. PCFs have been of interest to mathematicians for centuries and still are today, e.g. William Broucker's $\pi$ representation \cite{BrounckerContFrac}, Zudilin's work on difference equations and Catalan's constant (e.g., \cite{zudilin2003apery}), and other examples \cite{finch2004reviews,pickett2008another,lu2015some, SinghLevrie2018}. More on PCFs in Appendix Section \ref{app:ConjProofs}.

We demonstrate our approach by finding identities between a PCF and a fundamental constant substituted into a rational function. For efficient enumeration and expression aesthetics, we limit ourselves to integer polynomials on both sides of the equality. We propose two search algorithms: The first algorithm uses a Meet-In-The-Middle (MITM) technique, first executed to a relatively small precision to reduce the search space and eliminate mismatches. We then increase its precision with a higher number of PCF iterations on the remaining hits to validate them as conjectured RFs, and is therefore called MITM-RF. The second algorithm uses an optimization-based method, which we call Descent\&Repel, converging to integer lattice points that define conjectured RFs.

\newpage

Our MITM-RF algorithm was able to produce several novel conjectures that have short proofs\footnote{Several of the proofs were suggested by the community after our work was first put on arXiv. See Appendix Section \ref{app:ConjProofs} for more information.}, e.g.

\FloatBarrier
\begin{eqfloat}
\begin{equation}
\begin{split}
         \frac{4}{3\pi-8} = 3 - \frac{1\cdot1}{6 - \frac{2\cdot3}{9 - \frac{3\cdot5}{12 - \frac{4\cdot7}{..}}}} \\\\
         \frac{2}{\pi + 2} = 0 - \frac{1\cdot(3-2\cdot1)}{3 - \frac{2\cdot(3-2\cdot2)}{6 - \frac{3\cdot(3-2\cdot3)}{9 - \frac{4\cdot(3-2\cdot4)}{..}}}} \\
\end{split}
\quad\quad\quad
    \begin{split}
         \frac{e}{e-2} = 4 - \frac{1}{5 - \frac{2}{6 - \frac{3}{7 - \frac{4}{..}}}} \\\\
         \frac{1}{e - 2} = 1 + \frac{1}{1 + \frac{-1}{1 + \frac{2}{1 + \frac{-1}{1 + \frac{3}{..}}}}} \\
    \end{split}
\end{equation}
   \caption{A sample of automatically generated conjectures for mathematical formulas of fundamental constants, as generated by our proposed Ramanujan Machine by applying the MITM-RF algorithm. These conjectures were proven by contributions from the community following the publication in arXiv. Both results for $\pi$ converge exponentially and both results for $e$ converge super-exponentially. See Table \ref{table:ResTable_e} in the Appendix Section \ref{app:MoreRes} for additional results from our algorithms along with their convergence rates, which we separate to known formulas and to unknown formulas (to the best of our knowledge).
    \label{eq:conjectures1}}
\end{eqfloat}
\FloatBarrier

Our MITM-RF algorithm also produced novel conjectures that are currently still unproven: 
\FloatBarrier
\begin{eqfloat}
\begin{equation}
\begin{split}
         \frac{8}{\pi^2} = 1 - \frac{2\cdot1^4-1^3}{7 - \frac{2\cdot2^4-2^3}{19 - \frac{2\cdot3^4-3^3}{37 - \frac{2\cdot4^4-4^3}{..}}}} \\\\
         \frac{16}{4 + \pi^2} = 1 - \frac{2\cdot1^4-3\cdot1^3}{7 - \frac{2\cdot2^4-3\cdot2^3}{19 - \frac{2\cdot3^4-3\cdot3^3}{37 - \frac{2\cdot4^4-3\cdot4^3}{..}}}} \\
\end{split}
\quad\quad\quad
    \begin{split}
         \frac{12}{7 \zeta\left(3\right)} = 1\cdot2 - \frac{16\cdot 1^6}{3\cdot12 - \frac{16\cdot 2^6}{5\cdot32 - \frac{16\cdot 3^6}{7\cdot62 - \frac{16\cdot 4^6}{..}}}} \\\\
         \frac{2}{-1 + 2G} = 3 + 0\cdot7 - \frac{6\cdot1^3}{3 + 1\cdot10 - \frac{8\cdot2^3}{3 + 2\cdot13 - \frac{10\cdot3^3}{3 + 3\cdot16 - \frac{12\cdot4^3}{..}}}} \\
    \end{split}
\end{equation}
    \caption{To the best of our knowledge, these results are previously-unknown conjectures. $\zeta$ refers to the Riemann zeta function, and $G$ refers to the Catalan constant. The result for $e$ converges super-exponentially while the rest converge exponentially. See appendix section \ref{app:MoreRes} for additional results.
    \label{eq:conjectures2}}
\end{eqfloat}
\FloatBarrier
One may wonder whether the conjectures discovered by this work are indeed mathematical identities or merely mathematical coincidences that break down once enough digits are calculated. However, the method employed in this work makes it fairly unlikely for the conjectures to break down. For example,  the probability of finding a random match for an enumeration space of $10^{9}$ and result accuracy of more than $50$ digits, is smaller than $10^{-40}$. Our algorithms tested the conjectures for up to 2000 digits of accuracy. Nevertheless, such an accuracy does not replace the need of a formal proof\footnote{Examples of remarkable mathematical coincidences that hold true to high degree of approximation\cite{press} show that high accuracy will never substitute a formal proof.}.
We believe that many (if not all) of the new conjectures are indeed truths awaiting a rigorous proof, not only because of the accuracy used for the algorithms, but also due to the aesthetic nature of the conjectures. Indeed, since the original version of the paper and the project's website went online, many of the conjectures have been proven by colleagues in the wider mathematics community (e.g. \cite{ZeilbergerDougherty-Bliss2020}). We believe and hope that proofs of new computer-generated conjectures on fundamental constants will help create new mathematical knowledge and lead to new discoveries in the future.

\begin{figure}[h]
\begin{center}
\includegraphics[width=\textwidth]{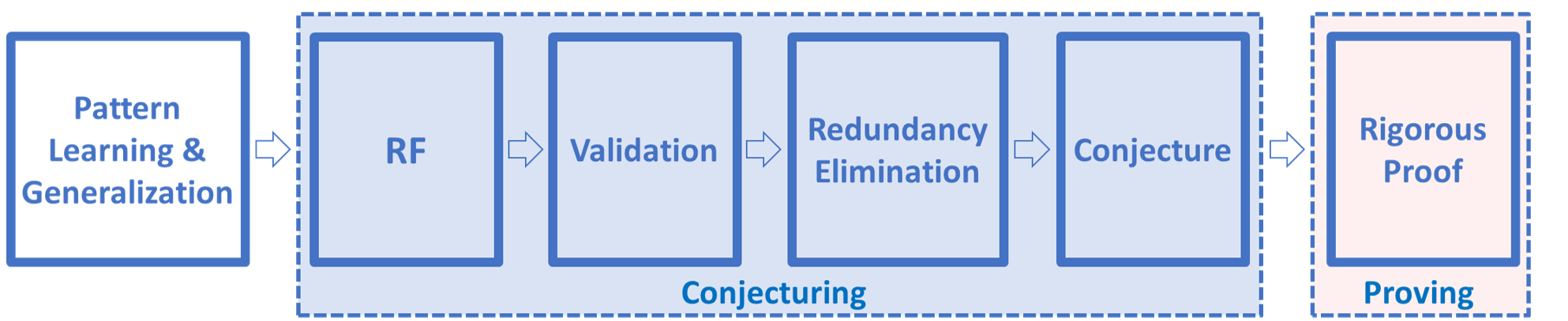}
\end{center}
\caption{\textbf{Conceptual flow of the wider concept of the Ramanujan Machine}. First, through various approaches of pattern learning \& generalization (Section \ref{sec:HypoDis}), we can generate a space of RF conjectures, e.g. PCFs. We then apply a search algorithm, validate potential conjectures, and remove redundant results. Finally, the validated results form mathematical conjectures that need to be proven analytically, thus closing a complete research endeavor from pattern generation to proof, and potentially new mathematical insight.}
\label{fig:AlgoPipeline}
\end{figure}

In contrast to the method we present, many known RFs of fundamental constants were discovered by conventional proofs, i.e., sequential logical steps derived from known properties of these constants. For example, several RFs of $e$ and $\pi$ were generated using the Taylor expansion of the exponent and the trigonometric functions and using Euler's continued fraction formula \cite{EulerContFrac}, which connects an infinite sum and an infinite PCF. In our work, we aim to reverse this process, finding new RFs for the fundamental constants using their numerical data alone, without any prior knowledge about their mathematical structure (Fig. \ref{fig:AlgoPipeline}). Each RF may enable reverse-engineering of the mathematical structure that produces it. In certain cases where the proof uses new techniques, it may also provide new insight on the field. Our approach is especially powerful in cases of empirical constants, such as the Feigenbaum constant from chaos theory (Table \ref{table:MoreCons}), which are derived numerically from simulations and have no analytic representation.

Given the success of our approach to finding new RFs for fundamental constants, there are many additional avenues for more advanced algorithms and future research. Inspired by worldwide collaborative efforts in mathematics such as the Great Internet Mersenne Prime Search (GIMPS), we launch the initiative \url{www.RamanujanMachine.com}, dedicated to finding new RFs for fundamental constants. The general community can donate computational time to find RFs, propose mathematical proofs for conjectured RFs, or suggest new algorithms for finding them (Appendix Section \ref{app:colab}). Since its inception, the Ramanujan Machine initiative has already yielded fruit, and several of the conjectures posed by our algorithms have already been proven (Appendix Section \ref{app:ConjProofs}).
\FloatBarrier
\section{Related Work}



The process of mathematical research is complex, nonlinear, and often leverages abstract mathematical intuition, all of which are difficult to express and study thoroughly. Respecting this fact, one may think in \emph{an oversimplified manner} about mathematical research as being separated into two main steps: conjecturing and proving (as in Fig. \ref{fig:AlgoPipeline}). 

While both steps received some attention in the literature, it is the \textit{second step} that was studied more extensively in the computer science literature and is known as Automated Theorem Proving (ATP), which focuses on proving existing conjectures. In ATP, algorithms already proved many theorems \cite{petkovvsek1996b}, such as the Four Color Theorem \cite{appel1989every}, the Robbins' problem (determining whether all Robbins algebras are Boolean algebras) \cite{mccune1997solution}, the Lorenz attractor problem \cite{tucker1999lorenz}, the Kepler Conjecture on the density of sphere packing \cite{hales2005proof}, as well as proving a conjectured identity for $\zeta(4)$ \cite{bailey1994experimental}, and other recent results \cite{chen2016automated,bansal2019holist,johansson2014hipster}, including applications of machine learning for ATP \cite{kaliszyk2014learning,urban2007malarea}.

Our work focuses on automating the \textit{first step} of the process, generation of new conjectures, which we refer to as Automated Conjecture Generation (ACG). Early work on ACG started with \cite{wang1960toward} and included substantial contributions such as the Automated Mathematician and EURISKO \cite{lenat1984and,lenat1982nature,lenat1982knowledge}, which envisioned the use of computers for the entire process of scientific discovery. Notable work by Fajtlowicz (called GRAFFITI) has found new conjectures in graph theory and matrix theory \cite{fajtlowicz1988conjectures}, by analyzing properties such as chromatic index and independence number on a large number of graphs and deducing general rules. Other works on the topic \cite{larson2005survey,larson2016automated,gauthier2016initial, colton2002hr, langley1983rediscovering,stoutemyer2017askconstants} used heuristics to discover new mathematical or physical concepts, rules, inequalities, or statements that were applied in a wide range of fields of mathematics (e.g., number theory) and of various natural sciences. ACG has also appeared as part of a combined approach supporting efforts in ATP \cite{buchberger2006theorema}. A particularly noteworthy algorithm in this context is PSLQ \cite{ferguson1999analysis}, which has been used to find formulas "by a combination of inspired guessing and extensive searching" \cite{bailey1997rapid}. Another inspiring example is a collaboration by Zeilberger and Zudilin on the irrationality measure of $\pi$ \cite{ZeilbergerZudilin2019}.

Our work differs from the others in a few manners. Namely, we present an end-to-end ACG, allowing for a fully-automatic process, including redundancy and false-positive removal without user input.
This way, we can validate our conjectures to arbitrary precision using numerical data as the ground truth for conjecturing. Most importantly, our conjectures focus on formulas for \textit{fundamental constants}. 


Proposing conjectures is sometimes more significant than proving them. For this reason, some of the most original mathematicians and scientists are known for their famous unsolved conjectures rather than for their solutions to other problems, like Fermat's last theorem, Hilbert's problems, Landau's problems, Hardy-Littlewood prime tuple conjecture, Birch-Swinnerton-Dyer conjecture, and of course the Riemann Hypothesis \cite{wiles1995modular,smale1998mathematical,hardy1923some,tate1965conjectures,landau1927vorlesungen}.
Maybe the most famous example is Ramanujan, who posed dozens of conjectures involving fundamental constants and considered them to be revelations from one of his goddesses \cite{berndt2012ramanujan}. In our work, \textbf{we aim to automate the process of conjectures generation and demonstrate it by providing new conjectures \emph{for fundamental constants}}.

By analyzing mathematical relationships of fundamental constants that are aesthetic and concise, the Ramanujan Machine can eventually extend known works of great mathematicians such as Gauss, Riemann, and Ramanujan himself to help us discover new mathematics.


\FloatBarrier
\section{The Meet-In-The-Middle-RF Algorithm}\label{sec:MITM}

The first algorithm we present searches for a PCF of a given a fundamental constant $c$ (e.g. $c=\pi$) in the following form:
\begin{equation}\label{eq:PCF}
    \frac{\gamma(c)}{\delta(c)} = f_i\left(\mathrm{PCF}(\alpha,\beta)\right)
\end{equation}
for a set of four integer polynomials ($\alpha$,$\beta$,$\gamma$,$\delta$), and a given set of functions $\{f_i\}$ (e.g. $f_1(x)=x,\ f_2(x) = \frac{1}{x},\,\dots$). $\mathrm{PCF}(\alpha,\beta)$ means the polynomial continued fraction with the partial numerator and denominator $a_{n} = \alpha(n) \, ; \, b_{n} = \beta(n)$, respectively, as defined in Eq. (\ref{PCFExample}).

\begin{figure}[h]
\begin{center}
\includegraphics[width=\textwidth]{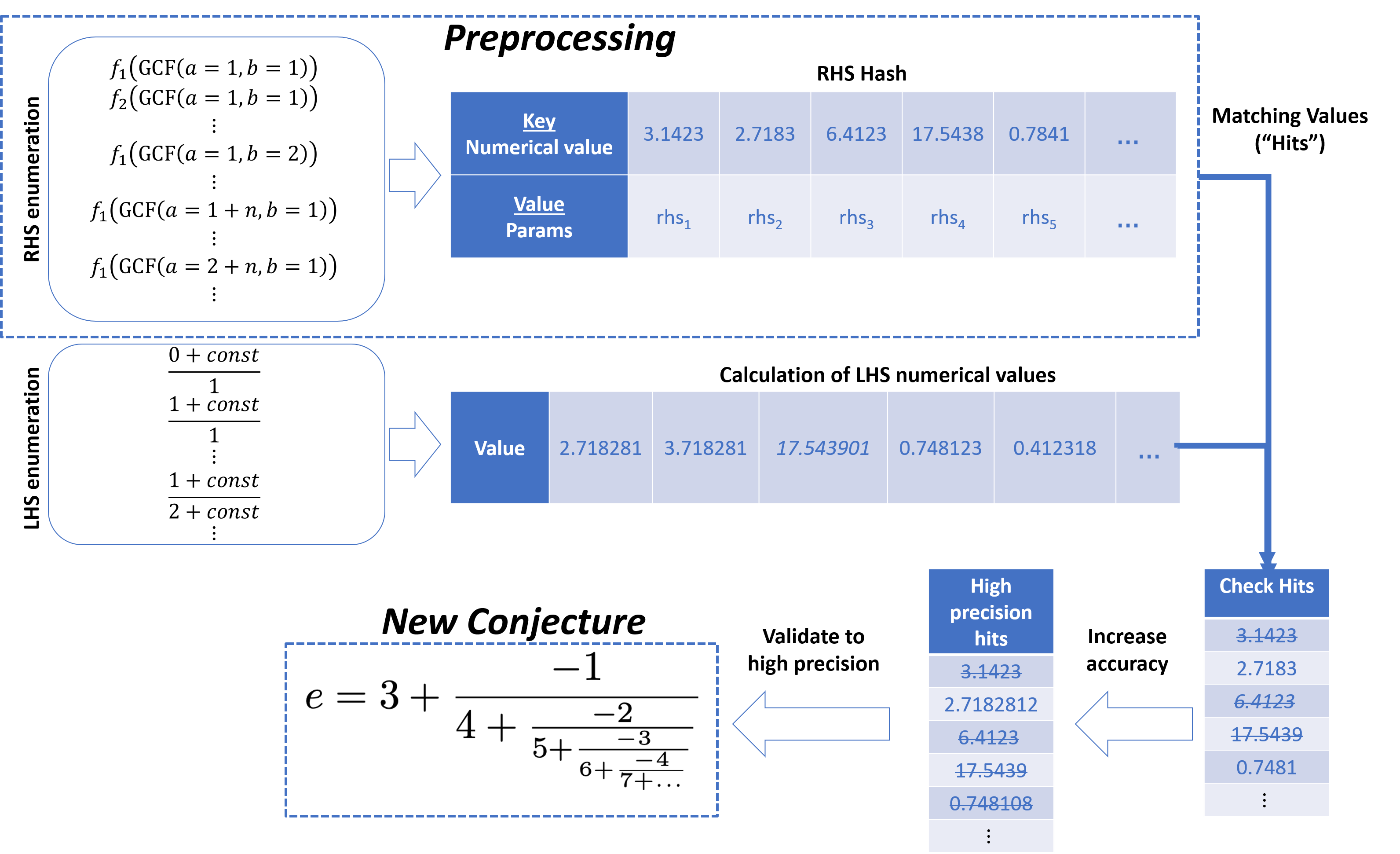}
\end{center}
\caption{\textbf{The Meet-In-The-Middle (MITM-RF) method}: first, we enumerate RHS to a low precision ($10$ digits), values are stored in a hash-table. Second, we enumerate over the LHS at low precision and search for matches. Finally, the matches are reevaluated to higher precision and compared again. The process can be repeated until reaching the required decimal precision, thus reducing false positives. The final results are then posed as new conjectures.}
\label{fig:MITMIIllustration}
\end{figure}

As showcased in Fig. \ref{fig:MITMIIllustration}, we start by enumerating over the two sides of Eq. \eqref{eq:PCF} and successively generating many different integer polynomials for $\alpha,\beta,\gamma,\delta$\footnote{We delete instances which produce trivial results like $\gamma=3\cdot\delta$ or $\alpha=0$ and instances whose $\beta$ polynomial has roots at natural numbers, which result in a finite PCF, necessarily representing a rational number}. We calculate the Right-Hand-Side (RHS) of each instance up to a limited number of iterations and store the results up to a pre-selected decimal point in a hash-table. The RHS is calculated with arbitrary-size integers, directly using the recurrence formula for the numerators $p_n$ and the denominators $q_n$ of the rational approximation of the PCF:
\begin{equation}
  \begin{split}
    q_{-1} &= 0\\
    q_{0} &= 1\\
    q_{n+1} &= a_{n+1}q_n + b_{n+1}q_{n-1}
  \end{split}
\quad\quad
  \begin{split}
    p_{-1} &= 1\\
    p_{0} &= a_0\\
    p_{n+1} &= a_{n+1}p_n + b_{n+1}p_{n-1}
  \end{split}
\end{equation}

We continue by evaluating the Left-Hand-Side (LHS) and attempt to locate each result in the hash-table, where successful attempts are considered as candidate solutions and will be referred to as "hits".
Since the LHS and RHS calculations are performed up to a limited precision, several of the hits are bound to be false positives. We then eliminate these false positives by calculating the RHS and LHS to higher precision (Fig. \ref{fig:MITMIIllustration}). 

A naive enumeration method is very computationally intensive with time complexity of $O(MN)$, where $M$ and $N$ are the LHS and RHS space size, respectively, and space complexity of $O(1)$. In our algorithm, we store the RHS in the hash-table in order to significantly reduce computation time at the expense of space. This makes the algorithm's time complexity $O(M+N)$ and its space complexity $O(N)$. Moreover, the hash-table of the PCF (RHS) can be saved and reused for further LHS enumerations, reducing future enumeration durations.

We also generalize the aforementioned algorithm to allow for $\alpha$ and $\beta$ to be interlaced sequences, i.e., they may consist of multiple integer polynomials. In a simple example of such a sequence, even values of $n$ are equal to one polynomial, and odd values of $n$ are equal to a different polynomial. Another improvement to our MITM algorithm is to swap the hash table content, i.e., store the LHS numerical values instead of the RHS values. This improvement enabled us to scale up our search and find additional PCF conjectures, for instance, the $\zeta(3)$ (i.e., Apéry's constant's constant) conjecture and the Catalan's constant conjecture presented in Eq. [\ref{eq:conjectures2}]. For the full implementation of our MITM-RF algorithm, see code on $www.RamanujanMachine.com$.

Our proposed MITM algorithm discovered new PCFs other than those previously known. Seeing how successful our algorithm was despite being relatively simplistic, we believe there is still ample room for new results. By leveraging more sophisticated algorithms, other results will follow, thus discovering hidden truths about even more fundamental constants, perhaps with formulas that are more complex than the PCFs used in this work (see Appendix Section \ref{app:colab}). 

After discovering dozens of PCFs, we empirically observed a relationship between the ratio of the polynomial order of $a_{n}$ and $b_{n}$, and the convergence rate of the formula. This relationship is proven in the Appendix Section \ref{app:proof} (Fig. \ref{fig:MITIMErr}).

\begin{figure}[h]
\begin{center}
\includegraphics[width=\textwidth]{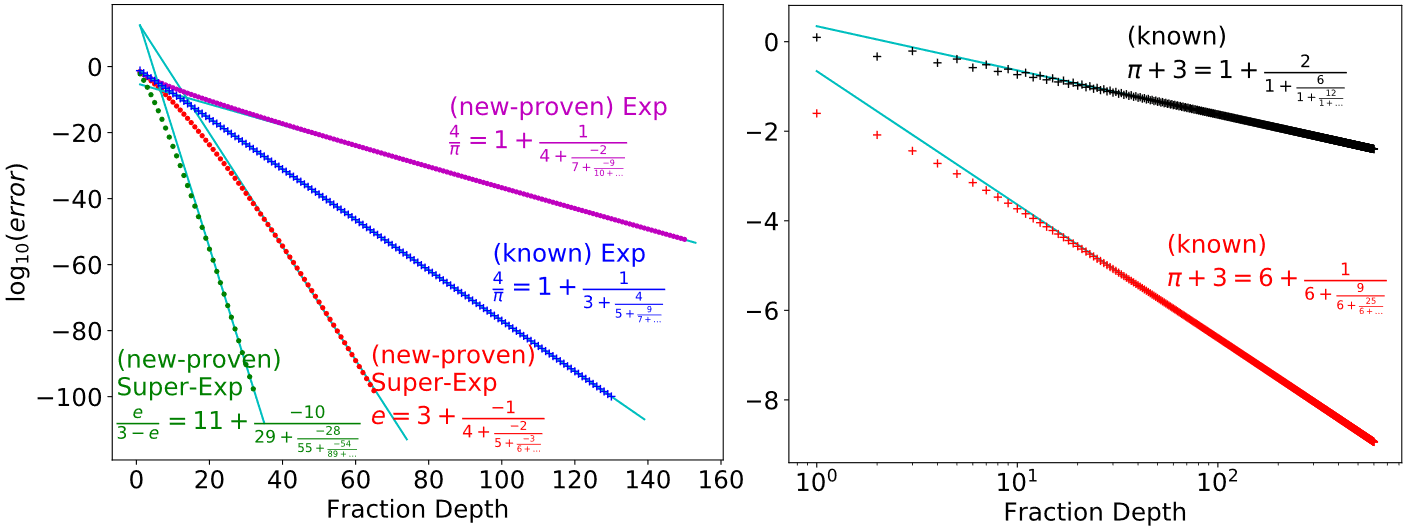}
\end{center}
\caption{\textbf{Convergence rates of the PCFs}. Plots of the absolute difference between the PCFs and the corresponding fundamental constant (i.e. the error) vs. the depth of the PCF. Every result found by our MITM-RF algorithm is tested numerically to 100 digits. On the left are PCFs that converge exponentially/super-exponentially  (validated numerically), and on the right are PCFs that converge polynomially. The vast majority of previously known PCFs for $\pi$ converge polynomially, while all of our newly found results converge exponentially.}
\label{fig:MITIMErr}
\end{figure}

\FloatBarrier
\section{Descent\&Repel}\label{sec:DesAndRepel}

\begin{figure}[h]
\begin{center}
\makebox[\textwidth][c]{\includegraphics[width=1\textwidth]{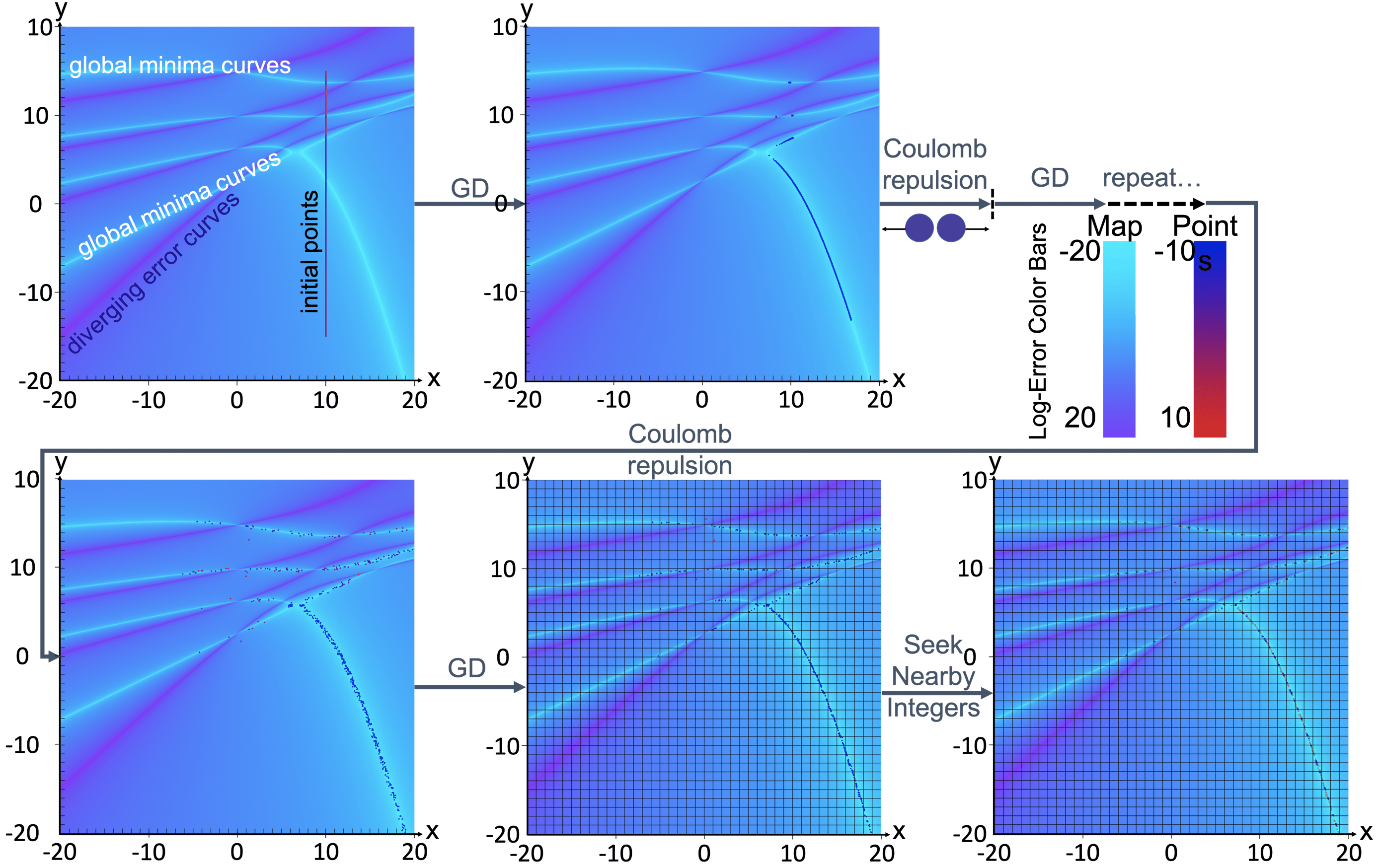}}%
\end{center}
\caption{\textbf{Schematic diagram of our Descent\&Repel algorithm for finding new RFs for fundamental constants, by relying on GD optimization}. The key observation that enables this method is that all minima are global ($\mathcal{L} = 0$) and appear as ($d-1$)-dimensional manifolds, where $d$ is the number of optimization variables. Starting with our initial conditions (in this example, consisting of 600 points on a vertical line), we perform ordinary GD alternated with "Coulomb" repulsion between all the points. Finally, to arrive to grid points, we perform GD toward integer points and toward the minimum curves, alternately. Lastly, we check whether any point satisfies the equation.}
\label{fig:GDyahel}
\end{figure}

We propose a GD optimization method and demonstrate its success in finding RFs. Although proven successful, the MITM-RF method is not trivially scalable. This issue can be targeted by either a more sophisticated variant or by switching to an optimization-based method, as is done by the following algorithm (Fig. \ref{fig:GDyahel}).

As explained in Section \ref{sec:MITM}, we want to find integer solutions to Eq. \eqref{eq:PCF}. This can also be written as the following constrained optimization problem:


\begin{equation}
    \underset{\alpha,\beta,\gamma,\delta}{\text{minimize}}\, \mathcal{L} = \left\lVert \frac{\gamma(\pi)}{\delta(\pi)} - \mathrm{PCF}(\alpha,\beta)\right\rVert \quad  \text{s.t.} \quad  \{\alpha,\beta,\gamma,\delta\}\ \subset  \mathbb{Z}[x]
\label{eq:minimize_gd}
\end{equation}
Solving this optimization problem with GD appears implausible since we are only satisfied with zero-error integer solutions. However, we found a significant feature of the loss landscape of the described problem that helped us develop a slightly modified GD, which we name 'Descent\&Repel' (Fig. \ref{fig:GDyahel}. Example of the results appear in Table \ref{table:newGDConst}). Without the restriction of being integers, minima are not $0$-dimensional points but rather ($d-1$)-dimensional manifolds with $d$ being the number of optimization variables, as would be expected given the single constraint. Moreover, we empirically observed that all minima are global, and their errors are zero (i.e., resulting in exact equality). Therefore, any GD process will result in a solution with $\mathcal{L} = 0$. It is well known that any real number can be expressed as a simple continued fraction \cite{jones1982survey} and the aforementioned feature hints that this may also be true for PCFs with integer polynomials.

\begin{table*}[h]
\begin{center}
\renewcommand{\arraystretch}{1.0}
\begin{tabular}{|l|l|l|l|}
\specialrule{.1em}{.05em}{.05em} 
\textbf{Convergence} & \textbf{Known} / \textbf{New} & \textbf{Formula} & \textbf{Polynomials} \\
\specialrule{.1em}{.05em}{.05em} 
    Exponential
            & known & $\frac{4}{\pi} = 1 + \frac{1}{3+\frac{4}{5+\frac{9}{7 +\ldots}}}$ & $a_n=1+2n,\ b_n=n^{2}$ \\ \cline{3-4}
\specialrule{.1em}{.05em}{.05em} 
    Super-Exponential
            & new and proven & $e = 3 + \frac{-1}{4+\frac{-2}{5+\frac{-3}{6+\ldots}}}$ & $a_n=3+n,\ b_n=-n$ \\ \cline{3-4}
\specialrule{.1em}{.05em}{.05em} 
\end{tabular}
\end{center}
\caption{RFs for $\pi$ and $e$ found in a proof-of-concept run of the Descent\&Repel algorithm.}
\label{table:newGDConst}
\end{table*}

We chose the variables of the optimization problem as the coefficients of the $\alpha,\beta,\gamma,\delta$ polynomials in Eq.\eqref{eq:minimize_gd}. The algorithm is initialized with a large set of points. In the specific examples we present, all initial conditions were set on a line, as is showcased in Fig. \ref{fig:GDyahel}.
The algorithm is then constructed of three main stages: GD, 'Repel', and Lattice GD. We iterate between the first two stages, and then perform the third stage once to converge to a possible result.
\begin{enumerate}
    \item (GD) We perform GD for all points, with $\mathcal{L}= \left\lVert \frac{\gamma(\pi)}{\delta(\pi)} - \mathrm{PCF}(\alpha,\beta)\right\rVert$, thus, $x_t = x_{t-1} - \mu\partial L$.
    \item ('Repel') We force all the points to push off one another via a Coulomb-like repulsion, $\frac{C}{\left\lVert a-b \right\rVert ^2}$. The 'repel' mechanism is used to increase the search space and thus the probability of finding a match in space.
    \item (Lattice GD) We enforce the constraint of integer results by alternating the optimization between the original loss Eq.\eqref{eq:minimize_gd} and an integer loss term that scales like the square of the difference between the value and its closest integer (round). This method allows us to find only points that have $\mathcal{L}=0$ on both losses, meaning an integer solution to our optimization problem.
\end{enumerate}

\FloatBarrier
\newpage
\section{Discussion}\label{sec:Discussion}

\subsection{Correspondence with the Mathematical Community}

Following the appearance of the initial version of our work on arXiv \cite{raayoni2019ramanujan}, many people ran our algorithms, and a few even found new conjectures. Others responded with proofs to the new formulas found by our algorithms. In fact, over the span of a few months, proofs for all the formulas in the original manuscript were presented. This led us to expand our search with the MITM-RF algorithm and find more intriguing results such as CFs for $\zeta(3)$, $\pi^2$, and Catalan's constant, \textbf{some of them still unproven}.

This back-and-forth dynamics between algorithms and mathematicians is the essence of what we believe can be achieved with automatically generated conjectures of fundamental constants. A recent example of this successful correspondence being the work done by Zeilberger's group \cite{ZeilbergerDougherty-Bliss2020}, proving part of our conjectures and generalizing them (see Appendix Section \ref{sub:DaughertyProofs}).

Another outcome of this mathematics-algorithm correspondence led us to arrive at some aesthetic generalizations. One example is the following conjecture:
$$\forall z \in \mathbb{C}: \quad  1+\frac{1\cdot(2\cdot z-1)}{4+\frac{2\cdot(2\cdot z-3)}{7+\frac{3\cdot(2\cdot z-5)}{10+\frac{4\cdot(2\cdot z-7)}{13+..}}}} = \frac{2^{2\cdot z +1}}{\pi\binom{2\cdot z}{z}}$$
This conjecture was generalized using several automatically generated conjectures (specific integer values for z), which were all generated by the aforementioned algorithms. Like many other results involving $\pi$, it can be proven using generalized hypergeometric functions. The proof is quite straightforward, provided prior knowledge about identities involving ratios of generalized hypergeometric functions, and is brought in Appendix Section \ref{sub:proof-of-general-pi} along with other proofs and related information. It remains to be seen whether related methods would be able to prove the conjectures in our work listed in Appendix Section \ref{app:MoreRes} as unproven.

These results are brought here as an example of how automatically-generated conjectures can be generalized to a wider conjecture and later a proof. We believe that this process could be used more widely with future results of the Ramanujan Machine, making automatically-generated conjectures on fundamental constants a catalyst for mathematical research. For an extended discussion see Appendix Section \ref{app:colab}.

\subsection{Applications in Mathematical Research}
New RF conjectures could have intriguing applications. Fast converging PCFs and other identities are being utilized for efficient calculation of different constants; for example, one of the most efficient historical methods to compute $\pi$ was based on a formula by Ramanujan \cite{borwein1989ramanujan}. More generally, new RFs could help us calculate other constants faster, like the super-exponential convergence that was demonstrated above for $e$. Another potential application of new RFs is for proving intrinsic properties of fundamental constants. An example is Apéry's proof that $\zeta(3)$ is irrational, done by representing it as a PCF \cite{apery1979irrationalite}. His work led to similar proofs for other constants. Specifically, it could be intriguing to look for PCFs for values of the Riemann $\zeta$ at odd integers \cite{Zudilin2018OddZeta}, because such PCFs may help prove their irrationality.

Looking forward, we consider systematic ways for generating a space of candidate RF conjectures, generalizing beyond the examples that we explored above. To establish new candidate mathematical conjectures, we envision harvesting the scientific literature (e.g., arXiv.org containing over 1.5M papers) as was done in \cite{tshitoyan} and generalizing RFs with machine learning algorithms such as clustering methods. The rich dataset available online should provide a strong ground truth for candidate RFs, which can be explored using algorithms similar to the ones described in this work. Such an approach may discover many new mathematical conjectures that go far beyond PCFs and can be explored in future work.
\subsection{The Universality of Fundamental Constants}

Our work provides the groundwork for a far more comprehensive study into fundamental constants and their underlying mathematical structure. Our proposed algorithms and their extensions found PCFs for the constants $\pi$, $e$, Catalan's constant, and $\zeta(3)$. Table \ref{table:MoreCons} presents a selection of additional fundamental constants of particular interest to our approach. For part of them, e.g., the Feigenbaum constants, \textit{no PCF, or any RF are known}. We also list a few examples of constants with intrinsic connections to the theory of PCF. Potentially the most interesting constants for further research are the ones coming from other fields, like number theory (not so ironically, some of them are also named after Ramanujan) and various fields of physics. With such constants, any new RF can point to a new hidden connection between fields of science. With further improvements and new algorithms applied to the thousands of fundamental constants in the literature, we expect many new RFs to be found.

\FloatBarrier
\begin{table*}[h]
\begin{center}
\renewcommand{\arraystretch}{1.0}
\begin{tabular}{|l|l|l|}
\specialrule{.1em}{.05em}{.05em} 
\textbf{Field} & Name           & Decimal Expansion  \\
\specialrule{.1em}{.05em}{.05em} 
    Related to Continued Fractions 
            & Lévy's constant & $\gamma = 3.275822\dots $  \\ \cline{2-3}
            & Khinchin's constant & $K_{0} = 2.685452\dots $ \\ \cline{2-3}
\specialrule{.1em}{.05em}{.05em} 
        Chaos Theory
            & First Feigenbaum constant & $\delta = 4.669201\dots$  \\ \cline{2-3}
            & Second Feigenbaum constant & $\alpha = 2.502907\dots$  \\ \cline{2-3}
            & Laplace Limit & $\lambda = 0.662743\dots$  \\ \cline{2-3}
\specialrule{.1em}{.05em}{.05em} 
    Number Theory 
        & Twin Prime constant & $\Pi_{2} = 0.660161\dots$  \\ \cline{2-3}
        & Meissel – Mertens constant & $M = 0.261497\dots$  \\ \cline{2-3}
        & Landau–Ramanujan constant& $ \Lambda = 0.764223\dots$  \\ \cline{2-3}
\specialrule{.1em}{.05em}{.05em} 
    Combinatorics 
        & Euler–Mascheroni constant& $\gamma = 0.577215\dots$  \\ \cline{2-3}
        & Catalan's constant & $G = 0.915965\dots$  \\ \cline{2-3}
\specialrule{.1em}{.05em}{.05em} 
    \dots 
        & \dots & \dots  \\ \cline{2-3}
\specialrule{.1em}{.05em}{.05em} 
\end{tabular}
\end{center}
\caption{\textbf{A sample of fundamental constants from different fields, which are all relevant targets for our method.} A wider list is available in \cite{finch2004reviews} and \cite{WolframContFrac}. For all of these, new RF conjectures will point to deep underlying connections. \textbf{There are thousands of additional constants for which enough numerical data exists, and our method is applicable}. With further improvement in our suggested approaches, along with new algorithms provided by the community, we expect that more new expressions will be found. Note that some constants in the table like the Feigenbaum constants have no analytical expression what-so-ever, and so far can only be computed using numerical simulation. Therefore, having a RF for them will reveal a hidden truth not only about the constant but also about the entire field to which it relates.}
\label{table:MoreCons}
\end{table*}
\FloatBarrier
\bibliographystyle{unsrt}
\bibliography{bibfile}

\appendix

\include{appendix}

\end{document}

%% file: appendix.tex
\setcounter{figure}{4}
\setcounter{table}{2}


\section{Additional Results by the MITM-RF Algorithm}\label{app:MoreRes}
In this section, we show a sample of polynomial continued fractions (PCFs) that were all found by our MITM-RF algorithm, and are summarized here. Our MITM-RF algorithm was able to reproduce previously known and proven results, along with new, previously unknown \footnote{New to the best of our knowledge. Some of these results might have already been found and published or be simple generalizations of previously known results. Regardless of their novelty, the Ramanujan Machine produced these results without any prior knowledge.} ones. Some of these results have been proven by the general mathematics community since being announced in \url{www.RamanujanMachine.com}, while others are still regarded as conjectures. For each RF, we provide its convergence rate and polynomials. 

These results help emphasize the novelty of our work, which is the concept of computer-generated conjectures and the specific algorithms we developed. We distinguish these points of novelty from the novelty of the generated results, where some can be \textbf{known}: i.e., we have found this result (or an equivalent form of it) in the literature, and therefore it serves as a proof-of-concept for the Ramanujan Machine but is not considered new. Results could also be \textbf{new and unproven}: i.e., we have not found this result in the literature, we consider it as a new conjecture, until proven or until an equivalent form that is unknown to us is found. Finally, results can be \textbf{new and proven}: i.e., proven after our first publication on arXiv.
\FloatBarrier
\begin{table}[h!]
\begin{center}
\begin{adjustbox}{width=1\columnwidth}
\begin{tabular}{|l|l|l|l|}
\specialrule{.1em}{.05em}{.05em} 
\textbf{Novelty} & \textbf{Formula} & \textbf{Polynomials} & \textbf{Convergence [$\frac{digits}{term}$]} \\
\specialrule{.1em}{.05em}{.05em}
    new and proven & 
    $ \frac{1 + e}{-1 + e} = 2 + \frac{1}{6 + \frac{1}{10 + \frac{1}{14 + \frac{1}{..}}}} $ &
    $ \begin{aligned}[t]
        a_{n}&=4n+2,\ b_{n}=1
    \end{aligned}$ &
    5.4905 *\\
\cline{1-4}
    new and proven & 
    $  \frac{3}{3 - e} = 11 - \frac{10}{29 - \frac{28}{55 - \frac{54}{89 - \frac{88}{..}}}}  $ &
    $ \begin{aligned}[t]
        a_{n}&=2n(2n+7)+11,\ b_{n}= -2n(2n + 3)
    \end{aligned}$ &
    4.9048 *\\
\cline{1-4}
    new and proven & 
    $1+\frac{e}{e-2} = 5 - \frac{4}{19 - \frac{18}{41 - \frac{40}{71 - \frac{70}{..}}}} $ &
    $ \begin{aligned}[t]
        a_{n}&=2n(2n+5)+5,\ b_{n}= -2n(2n+1)+2
    \end{aligned}$ &
    4.9018 *\\
\cline{1-4}
    new and proven & 
    $ \frac{e}{-24 + 9 e} = 6 - \frac{1}{7 - \frac{2}{8 - \frac{3}{9 - \frac{4}{..}}}} $ &
    $ \begin{aligned}[t]
        a_{n}&=6+n,\ b_{n}=-n
    \end{aligned}$ &
    2.1756 *\\
\cline{1-4}
    new and proven & 
    $ \frac{e}{6 - 2 e} = 5 - \frac{1}{6 - \frac{2}{7 - \frac{3}{8 - \frac{4}{..}}}} $ &
    $ \begin{aligned}[t]
        a_{n}&=5+n,\ b_{n}=-n
    \end{aligned}$ &
    2.1698 *\\
\cline{1-4}
    new and proven & 
    $  \frac{1}{-16 + 6 e} = 3 + \frac{1}{4 + \frac{2}{5 + \frac{3}{6 + \frac{4}{..}}}} $ &
    $ \begin{aligned}[t]
        a_{n}&=3+n,\ b_{n}=n
    \end{aligned}$ &
    2.1695 *\\
\cline{1-4}
    new and unproven & 
    $   \frac{6 e}{-3 + 2 e} = 7 - \frac{4}{14 - \frac{20}{23 - \frac{54}{34 - \frac{112}{..}}}} $ &
    $ \begin{aligned}[t]
        a_{n}&=n(n+6)+7,\ b_{n}=-(n+3)n^2
    \end{aligned}$ &
    2.164 *\\
\cline{1-4}
    new and proven & 
    $  \frac{e}{-2 + e} = 4 - \frac{1}{5 - \frac{2}{6 - \frac{3}{7 - \frac{4}{..}}}}$ &
    $ \begin{aligned}[t]
        a_{n}&=4+n,\ b_{n}=-n
    \end{aligned}$ &
    2.164 *\\
\cline{1-4}
    new and proven & 
    $   \frac{1}{-5 + 2 e} = 2 + \frac{1}{3 + \frac{2}{4 + \frac{3}{5 + \frac{4}{..}}}}$ &
    $ \begin{aligned}[t]
        a_{n}&=2+n,\ b_{n}=n
    \end{aligned}$ &
    2.1638*\\
\cline{1-4}
    new and unproven & 
    $    \frac{3}{-10 + 4 e} = 3 + \frac{4}{8 + \frac{20}{15 + \frac{54}{24 + \frac{112}{..}}}}$ &
    $ \begin{aligned}[t]
        a_{n}&=(n+1)(n+3),\ b_{n}= -(n+3)n^2
    \end{aligned}$ &
    2.1638 *\\
\cline{1-4}
    new and proven & 
    $ e = 3 - \frac{1}{4 - \frac{2}{5 - \frac{3}{6 - \frac{4}{..}}}} $ &
    $ \begin{aligned}[t]
        a_{n}&=3+n,\ b_{n}=-n
    \end{aligned}$ &
    2.1581*\\
\cline{1-4}
    new and proven & 
    $  \frac{1}{-2 + e} = 1 + \frac{1}{2 + \frac{2}{3 + \frac{3}{4 + \frac{4}{..}}}} $ &
    $ \begin{aligned}[t]
        a_{n}&=1+n,\ b_{n}=n
    \end{aligned}$ &
    2.158 *\\
\cline{1-4}
    known & 
    $ \frac{1}{-1 + e} = \frac{1}{1 + \frac{2}{2 + \frac{3}{3 + \frac{4}{..}}}}$ &
    $ \begin{aligned}[t]
        a_{n}&=n,\ b_{n}=n
    \end{aligned}$ &
    2.1522 *\\
\cline{1-4}
    new and proven & 
    $ \frac{e}{-1 + e} = 2 - \frac{1}{3 - \frac{2}{4 - \frac{3}{5 - \frac{4}{..}}}}$ &
    $ \begin{aligned}[t]
        a_{n}&=2+n,\ b_{n}=-n
    \end{aligned}$ &
    2.1522 *\\
\cline{1-4}
    new and unproven & 
    $ \frac{4 e}{-1 + 2 e} = 3 - \frac{3}{7 - \frac{16}{13 - \frac{45}{21 - \frac{96}{..}}}}$ &
    $ \begin{aligned}[t]
        a_{n}&=n(n+3)+3,\ b_{n}=-(n+2)n^2
    \end{aligned}$ &
   2.1493 *\\
\cline{1-4}

\specialrule{.1em}{.05em}{.05em} 
\end{tabular}
\end{adjustbox}
\end{center}
\caption{\textbf{Conjectures of $e$.} The cases marked by * have super-exponential convergence rates. The given value then refers to the number of digits per term, averaged over 200 terms.}
\label{table:ResTable_e}
\end{table}

\begin{table}[h]
\begin{center}
\begin{adjustbox}{width=1\columnwidth}
\begin{tabular}{|l|l|l|l|}
\specialrule{.1em}{.05em}{.05em} 
\textbf{Novelty} & \textbf{Formula} & \textbf{Polynomials} & \textbf{Convergence [$\frac{digits}{term}$]} \\
\specialrule{.1em}{.05em}{.05em}
    new and proven & 
    $ \frac{8}{-8 + 3 \pi} = 5 + \frac{5}{7 + \frac{12}{9 + \frac{21}{11 + \frac{32}{..}}}} $ &
    $ \begin{aligned}[t]
        a_{n}&=5+2n,\ b_{n}=n(n+4)
    \end{aligned}$ &
    0.75795 \\
\cline{1-4}
    new and proven & 
    $ \frac{4}{-2 + \pi} = 3 + \frac{3}{5 + \frac{8}{7 + \frac{15}{9 + \frac{24}{..}}}} $ &
    $ \begin{aligned}[t]
        a_{n}&=3+2n,\ b_{n}=n(n+2)
    \end{aligned}$ &
    0.75791 \\
\cline{1-4}
    known & 
    $ \frac{4}{\pi} = 1 + \frac{1}{3 + \frac{4}{5 + \frac{9}{7 + \frac{16}{..}}}} $ &

    $ \begin{aligned}[t]
        a_{n}&=1+2n,\ b_{n}=n^2
    \end{aligned}$ &
    0.75789 \\
\cline{1-4}
    new and proven & 
    $ \frac{-4 + 3 \pi}{20 - 6 \pi} = 5 - \frac{2}{8 - \frac{9}{11 - \frac{20}{14 - \frac{35}{..}}}} $ &
    $ \begin{aligned}[t]
        a_{n}&=5+3n,\ b_{n}=-(n+1)(2n-1)
    \end{aligned}$ &
    0.31382 \\
\cline{1-4}
    new and proven & 
    $ \frac{4}{-8 + 3 \pi} = 3 - \frac{1}{6 - \frac{6}{9 - \frac{15}{12 - \frac{28}{..}}}} $ &
    $ \begin{aligned}[t]
        a_{n}&=3+3n,\ b_{n}=-n(2n-1)
    \end{aligned}$ &
    0.31103 \\
\cline{1-4}
    new and proven & 
    $ \frac{8}{-8 + 3 \pi} = 6 - \frac{3}{9 - \frac{12}{12 - \frac{25}{15 - \frac{42}{..}}}} $ &
    $ \begin{aligned}[t]
        a_{n}&=6+3n,\ b_{n}=-(n+2)(2n-1)
    \end{aligned}$ &
    0.31103	\\
\cline{1-4}
    new and proven & 
    $ \frac{\pi}{4 - \pi} = 4 - \frac{2}{7 - \frac{9}{10 - \frac{20}{13 - \frac{35}{..}}}} $ &
    $ \begin{aligned}[t]
        a_{n}&=4+3n,\ b_{n}=-(n+1)(2n-1)
    \end{aligned}$ &
    0.30818	\\
\cline{1-4}
    new and proven & 
    $ \frac{2}{10 - 3 \pi} = 4 - \frac{3}{7 - \frac{10}{10 - \frac{21}{13 - \frac{36}{..}}}} $ &
    $ \begin{aligned}[t]
        a_{n}&=4+3n,\ b_{n}=-n(2n+1)
    \end{aligned}$ &
    0.30815 \\
\cline{1-4}
    new and proven & 
    $ \frac{2 \pi + 8}{\pi} = 5 - \frac{3}{8 - \frac{12}{11 - \frac{25}{14 - \frac{42}{..}}}} $ &	
    $ \begin{aligned}[t]
        a_{n}&=5+3n,\ b_{n}=-(n+2)(2n-1)
    \end{aligned}$ &
    0.30536 \\
\cline{1-4}
    new and proven & 
    $ \frac{2}{-2 + \pi} = 2 - \frac{1}{5 - \frac{6}{8 - \frac{15}{11 - \frac{28}{..}}}} $ &
    $ \begin{aligned}[t]
        a_{n}&=2+3n,\ b_{n}=-n(2n-1)
    \end{aligned}$ &
    0.30532	\\
\cline{1-4}
    new and proven & 
    $ \frac{6}{-8 + 3 \pi} = 5 - \frac{5}{8 - \frac{14}{11 - \frac{27}{14 - \frac{44}{..}}}} $ &
    $ \begin{aligned}[t]
        a_{n}&=5+3n,\ b_{n}=-n(2n+3)
    \end{aligned}$ &
    0.3053 \\
\cline{1-4}
    new and proven & 
    $ 1 + \frac{\pi}{2} = 3 - \frac{2}{6 - \frac{9}{9 - \frac{20}{12 - \frac{35}{..}}}} $ &
    $ \begin{aligned}[t]
        a_{n}&=3+3n,\ b_{n}=-(n+1)(2n-1)
    \end{aligned}$ &
    0.30243	\\
\cline{1-4}
    new and proven & 
    $ \frac{2}{4 - \pi} = 3 - \frac{3}{6 - \frac{10}{9 - \frac{21}{12 - \frac{36}{..}}}} $ &
    $ \begin{aligned}[t]
        a_{n}&=3+3n,\ b_{n}=-n(2n+1)
    \end{aligned}$ &
    0.3024 \\
\cline{1-4}
    new and proven & 
    $ \frac{2}{\pi} = 1 - \frac{1}{4 - \frac{6}{7 - \frac{15}{10 - \frac{28}{..}}}} $ &
    $ \begin{aligned}[t]
        a_{n}&=1+3n,\ b_{n}=-n(2n-1)
    \end{aligned}$ &
    0.29949	\\
\cline{1-4}
    known &
    $\frac{4}{\pi} = 1+\frac{1^2}{2+\frac{3^2}{2+\frac{5^2}{2+\ldots}}}$ &
    $ \begin{aligned}[t]
        a_n=2,\ b_n=\left(2n-1\right)^2
    \end{aligned}$& 
    polynomial \\ 
\cline{1-4}
    known &
    $\pi + 3= 6+\frac{1^2}{6+\frac{3^2}{6+\frac{5^2}{6+\ldots}}}$&
    $ \begin{aligned}[t]
        a_n=6,\ b_n=b_n = (2n-1)^2
    \end{aligned}$& 
    polynomial \\ 
\cline{1-4}
    known &
    $\frac{2}{2-\pi} = 3+\frac{-2\cdot 3}{1+\frac{-1\cdot2}{3+\frac{-4\cdot     5}{1+\frac{-2\cdot3}{3+\dots}}}}$ &
    $ \begin{aligned}[t]
            a_{n_1} &= 3,\ b_{n_1} = -2n(2n+1)\\
            a_{n_2} &= 1,\ b_{n_2} =-n(n-1)
    \end{aligned}$ &
    polynomial\\ 
\cline{1-4}
    known & 
    $\frac{6}{\pi^2-6} = 1+\frac{1^2}{1+\frac{1\cdot2}{1+\frac{2^2}{1+\frac{2\cdot3}{1+\dots}}}}$ & 
    $ \begin{aligned}[t]
        a_{n_1}&=1,\ b_{n_1}=n^2\\
        a_{n_2}&=1,\ b_{n_2}=n\left( n+1\right)
    \end{aligned}$ &
    polynomial\\

\specialrule{.1em}{.05em}{.05em} 
\end{tabular}
\end{adjustbox}
\end{center}
\caption{\textbf{Conjectures of $\pi$.}}
\label{table:ResTable_pi}
\end{table}

\begin{table}[h]
\begin{center}
\begin{adjustbox}{width=1\columnwidth}
\begin{tabular}{|l|l|l|l|}
\specialrule{.1em}{.05em}{.05em} 
\textbf{Novelty} & \textbf{Formula} & \textbf{Polynomials} & \textbf{Convergence [$\frac{digits}{term}$]} \\
\specialrule{.1em}{.05em}{.05em}
    known& 
    $ \frac{30}{\pi^2} = 3 + \frac{1}{25 + \frac{16}{69 + \frac{81}{135 + \frac{256}{223 + \frac{625}{..}}}}} $ &
    $ \begin{aligned}[t]
        a_{n}&=11n(n+1)+3,\ b_{n}=n^4
    \end{aligned}$ &
    2.069	\\
\cline{1-4}
    new and unproven & 
    $  \frac{8}{\pi^2}= 1 - \frac{2\cdot1^4-1^3}{7 - \frac{2\cdot2^4-2^3}{19 - \frac{2\cdot3^4-3^3}{37 - \frac{2\cdot4^4-4^3}{..}}}} $ &
    $ \begin{aligned}[t]
        a_{n}&=3n(n+1)+1,\ b_{n}= -(2n-1)n^3
    \end{aligned}$ &
    0.30241	\\
\cline{1-4}
    new and unproven & 
    $\frac{16}{4 + \pi^2}= 1 - \frac{2\cdot1^4-3\cdot1^3}{7 - \frac{2\cdot2^4-3\cdot2^3}{19 - \frac{2\cdot3^4-3\cdot3^3}{37 - \frac{2\cdot4^4-3\cdot4^3}{..}}}}$&
    $ \begin{aligned}[t]
        a_{n}&=3n(n+1)+1,\ b_{n}=  -2n^4 + 3n^3
    \end{aligned}$ &
    0.3111	\\
\cline{1-4}
    new and unproven & 
    $ \frac{24}{\pi^2} = 2 + \frac{8\cdot1^4}{16 + \frac{8\cdot2^4}{44 + \frac{8\cdot3^4}{86 + \frac{8\cdot4^4}{142 + \frac{8\cdot5^4}{..}}}}}$&
    $ \begin{aligned}[t]
        a_{n}&=7n(n+1)+2,\ b_{n}=8n^4
    \end{aligned}$ &
    0.89405	\\
\cline{1-4}
    new and unproven & 
    $ \frac{18}{\pi^2} = 2 - \frac{4\cdot1^4-2\cdot1^3}{13 - \frac{4\cdot2^4-2\cdot2^3}{34 - \frac{4\cdot3^4-2\cdot3^3}{65 - \frac{4\cdot4^4-2\cdot4^3}{..}}}} $&
    $ \begin{aligned}[t]
        a_{n}&=n(5n+6)+2,\ b_{n}= -4n^4 + 2n^3
    \end{aligned}$ &
    0.60045	\\
\cline{1-4}
    new and unproven & 
    $\frac{16}{-4 + \pi^{2}} = 3 - \frac{3}{13 - \frac{48}{29 - \frac{225}{51 - \frac{672}{79 - \frac{1575}{..}}}}}$&
    $ \begin{aligned}[t]
        a_{n}&=n(3n+7)+3,\ b_{n}=  -(n + 1)^2 (n + 3)(2n + 1)
    \end{aligned}$ &
    0.3082	\\

\cline{1-4}
    new and unproven & 
    $\frac{32}{\pi^{2}} = 3 + \frac{3}{13 - \frac{16}{29 - \frac{135}{51 - \frac{480}{79 - \frac{1225}{..}}}}}$	&
    $ \begin{aligned}[t]
        a_{n}&=n(3n+7)+3,\ b_{n}= -n^2 (n + 2)  (2n - 3)	
    \end{aligned}$ &
    0.31674 \\
\cline{1-4}
    new and unproven & 
    $\frac{16}{-8 + \pi^{2}} = 9 - \frac{9}{23 - \frac{96}{43 - \frac{375}{69 - \frac{1008}{101 - \frac{2205}{..}}}}}$&
    $ \begin{aligned}[t]
        a_{n}&=n(3n + 11) + 9,\ b_{n}= -n(n + 2)^2  (2n - 1)
    \end{aligned}$ &
    0.31384 \\
\cline{1-4}
    new and unproven & 
$\frac{16}{12 - \pi^{2}} = 9 - \frac{27}{23 - \frac{160}{43 - \frac{525}{69 - \frac{1296}{101 - \frac{2695}{..}}}}}$&
    $ \begin{aligned}[t]
        a_{n}&=n(3n + 11) + 9,\ b_{n}= -n(n + 2)^2  (2n + 1)	
    \end{aligned}$ &
    0.3052 \\
\cline{1-4}
    new and unproven & 
$\frac{32}{32 - 3 \pi^{2}} = 15 - \frac{45}{33 - \frac{240}{57 - \frac{735}{87 - \frac{1728}{123 - \frac{3465}{..}}}}}$ &
    $ \begin{aligned}[t]
        a_{n}&=n(3n + 15) + 15	,\ b_{n}=-n(n + 2)(n + 4)(2n + 1)	
    \end{aligned}$ &
    0.311 \\
\cline{1-4}
    new and unproven & 
$\frac{16 + 3 \pi^{2}}{16 - \pi^{2}} = 7 + \frac{8}{19 - \frac{27}{37 - \frac{192}{61 - \frac{625}{91 - \frac{1512}{..}}}}}$&
    $ \begin{aligned}[t]
        a_{n}&=n(3n + 9) + 7,\ b_{n}= -(n + 1)^3(2n - 3)	
    \end{aligned}$ &
    0.31948 \\
\cline{1-4}
    new and unproven & 
$\frac{18}{-8 + \pi^{2}} = 10 - \frac{10}{29 - \frac{112}{58 - \frac{486}{97 - \frac{1408}{146 - \frac{3250}{..}}}}}$&
    $ \begin{aligned}[t]
        a_{n}&=n(5n + 14) + 10	,\ b_{n}=-2n^3 (2n + 3)	
    \end{aligned}$ &
    0.60629 \\
\cline{1-4}

\specialrule{.1em}{.05em}{.05em}

    known &
    $\frac{1}{\zeta(3)} =0^{3} + 1^{3} -\frac{1^6}{1^{3} + 2^{3}-\frac{2^{2}}{2^{3} + 3^{3}-\frac{3^6}{3^{3} + 4^{3}-\ldots}}}$ &
    $\begin{aligned}[t]
        a_{n} &= n^{3} + (n+1)^{3} ,\ b_{n} &= - n^{6}
    \end{aligned}$ &
    polynomial \\
\cline{1-4}
    known* & 
   $\frac{5}{2\zeta(3)} = 2+  \frac{2\cdot1^{5}\cdot1}{2+1\cdot3\cdot7 + \frac{2\cdot2^{5}\cdot3}{2+1\cdot4\cdot10 +\frac{2\cdot3^{5}\cdot5}{2+1\cdot5\cdot13+\ldots}}}$&
    $ \begin{aligned}[t]
        a_{n}&=2 + n(2+n)(4+3n),\ b_{n}= 4n^6 - 2n^5
    \end{aligned}$ &
    0.60342	\\
\cline{1-4}
    known & 
   $\frac{6}{\zeta\left(3\right)} = 5 - \frac{1}{117 - \frac{64}{535 - \frac{729}{1463 - \frac{4096}{..}}}} $&
    $ \begin{aligned}[t]
        a_{n}&=(2 n + 1) (17n(n+1)+ 5),\ b_{n}= -n^6
    \end{aligned}$ &
    3.0316	\\
\cline{1-4}
    new and unproven & 
   $ \frac{8}{7 \zeta\left(3\right)} = 1\cdot 1 - \frac{1^6}{3\cdot 7 - \frac{2^6}{5\cdot 19 - \frac{3^6}{7\cdot 37 - \frac{4^6}{..}}}} $&
    $ \begin{aligned}[t]
        a_{n}&=(2 n + 1) (3n(n+1)+ 1),\ b_{n}= -n^6
    \end{aligned}$ &
    1.5158	\\
\cline{1-4}
    new and unproven & 
   $\frac{12}{7 \zeta\left(3\right)} = 1\cdot2 - \frac{16\cdot 1^6}{3\cdot12 - \frac{16\cdot 2^6}{5\cdot32 - \frac{16\cdot 3^6}{7\cdot62 - \frac{16\cdot 4^6}{..}}}}$&
    $ \begin{aligned}[t]
        a_{n}&=(2 n + 1) (5n(n+1) + 2),\ b_{n}=  -16n^6
    \end{aligned}$ &
    0.59602	\\
\cline{1-4}

\specialrule{.1em}{.05em}{.05em} 
    known* & 
   $ \frac{6}{- \pi \operatorname{acosh}{\left(2 \right)} + 8 G} = 2 - \frac{2}{19 - \frac{108}{56 - \frac{750}{113 - \frac{2744}{..}}}}$&
    $ \begin{aligned}[t]
        a_{n}&=10n^2 + 7n + 2,\ b_{n}=  -(2n-1)^4 - (2n-1)^3
    \end{aligned}$ &
    0.60046	\\
\cline{1-4}
    new and unproven & 
    $\frac{2}{-1 + 2G} = 3 - \frac{6}{13 - \frac{64}{29 - \frac{270}{51 - \frac{768}{79 - \frac{1750}{..}}}}}$&
    $ \begin{aligned}[t]
        a_{n}&=	n (3 n + 7) + 3,\ b_{n}=  -2n^ 3(n + 2)
    \end{aligned}$ &
	0.30387	\\

\cline{1-4}
    new and unproven & 
    $\frac{4}{-5 + 6 G} = 9 - \frac{18}{23 - \frac{128}{43 - \frac{450}{69 - \frac{1152}{101 - \frac{2450}{..}}}}}$&
    $ \begin{aligned}[t]
        a_{n}&=	n(3n + 11) + 9,\ b_{n}=  -2n^2(n + 2)^ 2
    \end{aligned}$ &
	0.30958	\\

\cline{1-4}
    new and unproven & 
    $\frac{6}{17 - 18G} = 13 - \frac{32}{29 - \frac{180}{51 - \frac{576}{79 - \frac{1400}{113 - \frac{2880}{..}}}}}$&
    $ \begin{aligned}[t]
        a_{n}&=	n(3n + 13) + 13,\ b_{n}=-2 n(n + 1)^ 2(n+3)
    \end{aligned}$ &
	0.31239	\\

\cline{1-4}

\specialrule{.1em}{.05em}{.05em} 
\end{tabular}
\end{adjustbox}
\end{center}
\caption{\textbf{Conjectures of other constants.} Sample of conjectures to other mathematical constants including $\pi^2$, $\zeta(3)$(Apéry's constant) and $G$(Catalan's Constant). The results marked as \textbf{known*} in this table are ones that we derived from other expressions of these constants}
\label{table:ResTable_other}
\end{table}
\FloatBarrier
\newpage

Another family of results easily observed among the rest is:
$$\varphi^k = L_k - \frac{(-1)^k}{L_k - \frac{(-1)^k}{L_k - \frac{(-1)^k}{..}}}$$ $L_k$ being the $k^{\text{th}}$ Lucas number and $\varphi=\frac{1+\sqrt{5}}{2}$, the golden ratio.
The convergence rate for this conjecture increases monotonically with k. 
Although the proof is immediate (and is left as an exercise for the reader), we chose to note this conjecture because of its highly aesthetic nature.

\section{Collaborative Algorithm-Enhanced Mathematics}\label{app:colab}

In the most general sense, the Ramanujan Machine is a methodology that generates conjectures on fundamental constants. On average, the more computational power and time spent by the algorithm on a selected parameter space, the more conjectures it should generate (assuming that RFs exist in the selected parameter space). Since conjecturing is only one aspect of mathematical discovery, it is clear that proving the conjectures is necessary to confirm them as mathematical truths. This section argues how one may leverage these facts to enjoy network effects that inspire the wider community about mathematics and specifically number theory.

We created the Ramanujan Machine as an open-source project that is fully available to the community on \url{www.RamanujanMachine.com}. With our ongoing development, individuals around the world would be able to donate their computational power to the mission of discovering new mathematical structures and mathematical equations by downloading the Ramanujan Machine "screen saver". Similarly to SETI (Search for Extraterrestrial Intelligence), we plan to have the Ramanujan Machine algorithm distribute via BOINC the various computational tasks to every idle computer in the network. Given enough computers in the Ramanujan Machine network, the computational power can be many folds higher than the computational power used to generate the conjectures in this work thus far.

We believe this methodology can inspire the greater community about mathematics. In order to achieve this goal, the site \url{www.RamanujanMachine.com} is regularly updated with conjectures generated by our algorithms. When a specific computer in the network discovers a new conjecture, after verifying that the conjecture had not been discovered elsewhere, the owner of the laptop will receive the credit for contributing his or her computer power to discover the conjecture and the credit is maintained in a leadership board. This way, the Ramanujan Machine also encourages a wider audience who is intrigued by the world of mathematics to contribute to it. People who do not have much time but do have the computational power at hand can contribute the computational power alone and may thus generate conjectures. Others, who may be interested in getting involved with deeper mathematical details, may suggest proofs to the conjectures discovered and thus contribute new mathematics and validate the conjectures. A third group may include people who rather dive deeper into the algorithms, and may propose, modify, and contribute new algorithms to extend the reach and efficiency of the Ramanujan Machine in discovering new conjectures.

It is important to emphasize that the methodology introduced in this work, and specifically the last comment, can be expanded far beyond continued fractions, number theory or mathematics. The Ramanujan Machine is an example of a broader methodology (Fig. \ref{fig:AlgoPipeline}).

We conclude with an intriguing yet admittedly speculative comment. Historically, the intuition of geniuses was instrumental in pushing mathematical research forward. Examples include Gauss, Ramanujan, and Fermat, all of whom found special cases that ended up being generalized and which eventually opened entire fields of mathematics. One such case involves Gauss' famous statement (mentioned in the introduction): "I have the result but I do not yet know how to get it" \cite{asimov1988isaac}: In 1799, Gauss discovered the relation between the lemniscate sine function and the arithmetic-geometric mean iteration. This discovery was made by his observation of a numerical equality between his calculation and a value in Stirling's tables of integrals. We believe that a computer algorithm could have achieved such a discovery.

While algorithms have the advantage of sheer computer power, it is clear that they may be limited in terms of their capacity. However, it is worthy to emphasize that algorithms have other advantages that may allow them to go beyond human knowledge and intuition. We specifically refer to the fact that algorithms like the ones discussed in this work are detached from the approaches humans may take. This detachment can be an advantage that may lead to insights that are extremely unlikely to be discovered otherwise. As an example, consider Google's Alpha-Go machine \cite{silver2017mastering}, making game moves that make no sense to humans and would not have been played by any human, yet proved to be brilliant. Analogously here, while some automatic conjectures can be shown to be a direct result of known mathematical structures, others may not be. Either way, the computer did not use any such structure. Therefore, the same algorithms could just as well discover new conjectures for which no proof is currently known, or that is not derived from any known mathematical structure. In this way, the algorithms can contribute to existing mathematical knowledge. This may even be the case with our results listed as unproven (for a full list, see Appendix Section \ref{app:MoreRes}). We do not know whether these results will eventually be found easy to prove, but regardless, they demonstrate the potential of algorithms for automatic conjecturing.

\section{Outlook on Continued Fractions}\label{sec:HypoDis}
Many questions remain open regarding the nature of continued fractions. One such question is, which fundamental constants can even be expressed with PCFs? This outlook is thoroughly discussed in \cite{McLaughlinWyshinski2004}, which also points out that almost all real numbers do not have a PCF expansion since the set of all polynomial continued fractions is a countable set. With that being said, many algebraic and transcendental numbers do have a PCF expansion. It is then tempting to ask whether any computable number may have a PCF expansion (or another form of RF continued fraction).

It appears that for some numbers (e.g., $e$, $\pi$), numerous PCF representations were found with low polynomial orders, while for others only a few. This raises a question regarding each fundamental constant, does it have a limited number of PCF forms or rather an infinite family (excluding trivial degenerate cases)? How does the number of representations vary between different constants? The reason our algorithms only found a few representations for certain constants may be that other PCF representations of them require a high degree of polynomials or high coefficients that exceed the domain of our search algorithms.

Another important property of PCFs that we explore in this research is their rate of convergence (Fig. \ref{fig:MITIMErr}). We noticed and later proved that the rate of convergence is a function of the degrees of the $\alpha,\beta$ polynomials: When $\frac{\deg(\beta)}{\deg(\alpha)}>2$, then the convergence is polynomial in the PCF depth. When the ratio is smaller than $2$, then the convergence is super-exponential. When the ratio is precisely $2$, then the convergence can be exponential, depending on more subtle conditions (see Appendix Section \ref{app:proof} for details). This result allowed us to improve the MITM-RF algorithm further.

\newpage
\section{PCF Convergence Rate}\label{app:proof}
The method detailed in Section \ref{sec:MITM} requires estimating the expected accuracy from a finite approximation of PCFs. In this section, we characterize the convergence rate of the PCFs. Also, we describe a trick that improves this convergence rate for the exponential case.

For two sets of numbers $\left\{ a_{n}\right\} _{n=0}^{\infty},\ \left\{ b_{n}\right\} _{n=1}^{\infty}$,
we define the polynomial continued fraction (PCF) generated by them as
\begin{eqnarray*}
\left[a_{0};\left(b_{1},a_{1}\right),\left(b_{2},a_{2}\right),\ldots\right] & \coloneqq & a_{0}+\frac{b_{1}}{a_{1}+\frac{b_{2}}{a_{2}+\ldots}}
\end{eqnarray*}
and the partial PCF as
\begin{eqnarray*}
\eta_{n} & \coloneqq & \left[a_{0};\left(b_{1},a_{1}\right),\left(b_{2},a_{2}\right),\ldots,\left(b_{n},a_{n}\right)\right]
\end{eqnarray*}
If the limit exists, we define:
\begin{eqnarray*}
\eta & \coloneqq & \lim_{n\rightarrow\infty}\eta_{n}
\end{eqnarray*}
We also define the tail:
\begin{eqnarray*}
\tau_n & \coloneqq & \left[a_n;\left(b_{n+1},a_{n+1}\right),\left(b_{n+2},a_{n+2}\right),\ldots\right]
\end{eqnarray*}
From there it follows that:
\begin{eqnarray*}
\eta & = & \frac{\tilde{p}_n(\tau_n)}{\tilde{q}_n(\tau_n)}
\end{eqnarray*}
where $\tilde{p}_n,\tilde{q}_n\in\mathbb{F}_1\left[x\right]$ are polynomials of degree $1$
whose coefficients depend on $\{a_i\}_{i=0}^{n-1},\{b_i\}_{i=1}^n$. Specifically, for $a_i,b_i\in\mathbb{Z}$ we have $\tilde{p}_n,\tilde{q}_n\in \mathbb{Z}_1[x]$.

It was shown \cite{jones1982survey} that the partial PCF $\eta_{n}$ can be computed as a series of Matrix-Vector multiplications, with $p_{n}$ and $q_{n}$ being the numerator and denominator of $\eta_n$, respectively:
\begin{eqnarray*}
\left(\begin{matrix}p_{0}\\
p_{-1}
\end{matrix}\right) & \coloneqq & \left(\begin{matrix}a_{0}\\
1
\end{matrix}\right)\\
\left(\begin{matrix}q_{0}\\
q_{-1}
\end{matrix}\right) & \coloneqq & \left(\begin{matrix}1\\
0
\end{matrix}\right)\\
\left(\begin{matrix}
p_{n+1} & q_{n+1}\\
p_{n} & q_{n}
\end{matrix}\right) & = & \left(\begin{matrix}
a_{n} & b_{n}\\
1 & 0
\end{matrix}\right)\left(\begin{matrix}
p_{n} & q_{n}\\
p_{n-1} & q_{n-1}
\end{matrix}\right)\\
 & \Downarrow\\
p_{n+1} & = & a_{n}p_{n}+b_{n}p_{n-1}\\
q_{n+1} & = & a_{n}q_{n}+b_{n}q_{n-1}
\end{eqnarray*}
From which $\eta_{n}$ can be calculated using:
\begin{eqnarray*}
\eta_{n} & = & \frac{p_{n}}{q_{n}}
\end{eqnarray*}
One can also conclude from the above (using $b_0 \coloneqq 1$):
\begin{eqnarray}
\label{eq:pcf_recursive_matrices_calc}
\left(\begin{matrix}
p_{n+1} & q_{n+1}\\
p_{n} & q_{n}
\end{matrix}\right) & = & \prod_{i=0}^n\left(\begin{matrix}a_{n} & b_{n}\\
1 & 0
\end{matrix}\right)
\end{eqnarray}

In the following sections we discuss PCFs with integer polynomials for $a_n$ and $b_n$:
\begin{eqnarray*}
a\left(x\right),b\left(x\right) & \in & \mathbb{Z}\left[x\right]\\
a_{n} & = & a\left(n\right)\\
b_{n} & = & b\left(n\right)
\end{eqnarray*}
which we will abbreviate as PCFs.

\FloatBarrier
\subsection{\label{sub:PCF-Error-Bound}Error Bound on a Finite Calculation of a PCF}

To find the error of a finite calculation of a PCF, we can take the determinant of Eq. (\ref{eq:pcf_recursive_matrices_calc})
\begin{eqnarray*}
p_{n+1}q_{n}-q_{n+1}p_{n} & = & \left(-1\right)^{n}\prod_{i=1}^{n}b_{i}\\
 & \Downarrow\\
\eta_{n+1}-\eta_{n} & = & \left(-1\right)^{n}\frac{\prod_{i=1}^{n}b_{i}}{q_{n+1}q_{n}}
\end{eqnarray*}

We would like to show that the following is a Leibniz series:
\[\sum_{i=k}^{\infty}\eta_{i+1}-\eta_{i}\]
We prove it for $b_{i}$ that becomes positive for a large enough $i$ (there are ways to generalize this proof that we do not discuss here).
Since $a_i$ is a polynomial, there exist $k\in\mathbb{N}$ such that $a_i$ keeps a constant sign for all $i > k$. Now we can show that depending on the values of $a_i$ and $b_i$, the series $q_i$ either has a constant sign or an alternating sign. Therefore, $q_iq_{i+1}$ will have a constant sign for all $i>k$. This proves that the series $\eta_{i+1}-\eta_{i}$ is a Leibniz series.

To prove that $q_i$ will have either a constant sign or an alternating sign, we first note that converting $a_n \mapsto -a_n$ only changes the sign of the value of the continued fraction, but neither the magnitude of the error nor the convergence rate will change. Therefore, we assume WLOG that for all $i>k$, $a_i>0$. Now, assuming that there are two consecutive terms $q_i,q_{i+1}$ ($i>k$) with the same sign, then the rest of the series will remain with the same sign. We can see that by writing:
\begin{eqnarray*}
q_{i+1} = \underbrace{a_n}_{>0}  q_i +\underbrace{b_n}_{>0} q_{i-1} \geq q_i + q_{i-1} > 0
\end{eqnarray*}
in case that $q_i,q_{i-1}>0$, or
\begin{eqnarray*}
q_{i+1} = \underbrace{a_n}_{>0}  q_i +\underbrace{b_n}_{>0} q_{i-1} \leq q_i + q_{i-1} < 0
\end{eqnarray*}
if they are both negative. Therefore, in the case that there are two consecutive terms with the same sign, the claim holds. In the other case, by assumption, there are no two consecutive terms with the same sign, therefore the signs of the series $\{q_i\}_{i>k}$ alternate. In both cases, the multiplication $q_{i+1}q_i$ will have a constant sign for all $i>k$.

Using the convergence of a Leibniz series we get:
\begin{eqnarray*}
\eta_{n+1} & = & \eta_{k}+\sum_{i=k}^{n}\eta_{i+1}-\eta_{i}=\eta_{k}+\sum_{i=k}^{n}\left(-1\right)^{i}\frac{\prod_{j=1}^{i}b_{j}}{q_{i+1}q_{i}}
\end{eqnarray*}
Therefore, the following relation is achieved:
\begin{eqnarray*}
\forall n\geq k\quad\eta_{2n+\kappa} & \leq\lim_{n\rightarrow\infty}\eta_{2n+\kappa}\leq\lim_{n\rightarrow\infty}\eta_{2n+\kappa-1} & \leq\eta_{2n+\kappa-1}
\end{eqnarray*}
where $\kappa\in\left\{ 0,1\right\} $, depending on the sign of $\prod_{i=1}^{k}b_{i}$
and $k\mod2$. 

Hence we get that the PCF converges $\eta$ and the error can be bound by
\begin{eqnarray*}
\left|\eta-\eta_{n}\right| & \leq & \left|\frac{\prod_{i=1}^{n}b_{i}}{q_{n+1}q_{n}}\right|
\end{eqnarray*}

\FloatBarrier
\subsection{1-Periodic PCF}\label{sec:1Pred}

A PCF is called $k$-periodic if $\forall n\in\mathbb{N}\ a_{n}=a_{n+k},\ b_{n}=b_{n+k}.$
A $1$-periodic PCF is one of the form:
\[
a+\frac{b}{a+\frac{b}{a+\ldots}}
\]
hence for both $\left\{w_{n}\right\}=\left\{p_{n}\right\}$ or $\left\{w_{n}\right\}=\left\{q_{n}\right\}:$
\begin{eqnarray*}
\left(\begin{matrix}w_{n+1}\\
w_{n}
\end{matrix}\right) & \coloneqq & \underbrace{\left(\begin{matrix}a & b\\
1 & 0
\end{matrix}\right)}_{\begin{subarray}{c}
\text{promoter}\\
\text{matrix}
\end{subarray}}\left(\begin{matrix}w_{n}\\
w_{n-1}
\end{matrix}\right)
\end{eqnarray*}
For $a^{2}>-4b$, we find that the promoter matrix is real-diagonalizable
\begin{eqnarray*}
\mbox{eigvals}\left(\begin{matrix}a & b\\
1 & 0
\end{matrix}\right) & = & \left\{ \frac{a\pm\sqrt{a^{2}+4b}}{2}\right\} =\left\{ \lambda_{\pm}\right\} \\
\mbox{eigvecs}\left(\begin{matrix}a & b\\
1 & 0
\end{matrix}\right) & = & \left\{ \left(\begin{matrix}\lambda_{\pm}\\
1
\end{matrix}\right)\right\} =\left\{ \mathbf{v}_{\pm}\right\} \\
 & \Downarrow\\
\left(\begin{matrix}w_{0}\\
w_{1}
\end{matrix}\right) & = & \kappa_{+}\mathbf{v}_{+}+\kappa_{-}\mathbf{v}_{-}\\
\left(\begin{matrix}w_{n}\\
w_{n-1}
\end{matrix}\right) & = & \lambda_{+}^{n}\kappa_{+}\mathbf{v}_{+}+\lambda_{-}^{n}\kappa_{-}\mathbf{v}_{-}\\
\end{eqnarray*}
and from there, the decomposition for $p,q$ is:
\begin{eqnarray*}
\\
\left(\begin{matrix}p_{0}\\
p_{-1}
\end{matrix}\right) & = & \left(\begin{matrix}a_{0}\\
1
\end{matrix}\right)=\frac{\lambda_{+}}{\sqrt{a^{2}+4b}}\mathbf{v}_{+}-\frac{\lambda_{-}}{\sqrt{a^{2}+4b}}\mathbf{v}_{-}\\
\left(\begin{matrix}q_{0}\\
q_{-1}
\end{matrix}\right) & = & \left(\begin{matrix}1\\
0
\end{matrix}\right)=\frac{\mathbf{v}_{+}-\mathbf{v}_{-}}{\sqrt{a^{2}+4b}}
\end{eqnarray*}
Thus,
\begin{eqnarray*}
\frac{p_{n-1}}{q_{n-1}} & = & \frac{\frac{1}{\sqrt{a^{2}+4b}}\left(\lambda_{+}^{n+1}-\lambda_{-}^{n+1}\right)}{\frac{1}{\sqrt{a^{2}+4b}}\left(\lambda_{+}^{n}-\lambda_{-}^{n}\right)}=\frac{\lambda_{+}^{n+1}-\lambda_{-}^{n+1}}{\lambda_{+}^{n}-\lambda_{-}^{n}}
\end{eqnarray*}
For $a>0$ we get that $\left|\lambda_{+}\right|>\left|\lambda_{-}\right|\geq0$, and hence:
\begin{eqnarray*}
\frac{p_{n-1}}{q_{n-1}} & = & \lambda_{+}\frac{1-\left(\frac{\lambda_{-}}{\lambda_{+}}\right)^{n+1}}{1-\left(\frac{\lambda_{-}}{\lambda_{+}}\right)^{n}}\\
\lim_{n\rightarrow\infty}\eta_{n} & = & \lambda_{+}
\end{eqnarray*}
While in the case $a<0$, we have $\left|\lambda_{-}\right|>\left|\lambda_{+}\right|\geq0$, which in turn results in:
\begin{eqnarray*}
\lim_{n\rightarrow\infty}\eta_{n} & = & -\lambda_{-}
\end{eqnarray*}
Note that provided that the PCF converges  ($\exists\lim_{n\rightarrow\infty}\eta_{n}$),
then $\eta=a+\frac{b}{\eta}$, yielding a quadratic equation with
the same results.

\FloatBarrier
\subsection{Types of Convergence}
Not every continued fraction converges. In the case it does, its rate of convergence is either: exponential, super-exponential, or sub-exponential (which seems to be at a polynomial rate). When the continued fraction does not converge, it may oscillate between a set of values or ``converge'' to an oscillating cycle of a certain periodicity, meaning that for a $k$-oscillation with values $\left\{ o_{i}\right\} _{i=0}^{k-1}$, we have $\lim_{n\rightarrow\infty}\left|\eta_{n}-o_{n\mod k}\right|=0$.

In the following parts, we analyze the PCFs behaviour with regard to its defining polynomials $a,b$. We will use the following notation:
\begin{eqnarray*}
d_{a} & \coloneqq & \mbox{deg}\left(a\right)\\
d_{b} & \coloneqq & \mbox{deg}\left(b\right)\\
a\left(x\right) & = & \sum_{j=1}^{d_{a}}\alpha_{j}x^{j}\\
b\left(x\right) & = & \sum_{i=0}^{d_{b}}\beta_{j}x^{i}
\end{eqnarray*}

For a more accessible analysis of the PCFs behavior, we use the equivalence transformation and define its semi-canonical form as\footnote{This is well defined, as we are examining the tail's behavior. Therefore neglect $n$'s for
which $a_{n}=0$, as they are finite.}:
\begin{eqnarray*}
\forall n\in\mathbb{N}\ c_{n} & \coloneqq & \frac{b_{n}}{a_{n-1}a_{n}}\\
a_{0}\left(1+\frac{\frac{b_{1}}{a_{0}a_{1}}}{1+\frac{\frac{b_{2}}{a_{1}a_{2}}}{1+\ldots}}\right) & \eqqcolon & \left[a_{0};\left(c_{1},c_{2},\ldots\right)\right]
\end{eqnarray*}
From there it follows:
\begin{eqnarray*}
\eta_{n} & = & \left[a_{0};\left(c_{1},\ldots,c_{n}\right)\right]
\end{eqnarray*}
Unless stated otherwise, we will regard only the main part of the above PCF:
\begin{eqnarray*}
1+\frac{\frac{b_{1}}{a_{0}a_{1}}}{1+\frac{\frac{b_{2}}{a_{1}a_{2}}}{1+\ldots}} & = & 1+\frac{c_{1}}{1+\frac{c_{2}}{1+\ldots}}
\end{eqnarray*}

We now recognize $3$ distinct cases. In the first case, denoted as the exponential case, we have:
\begin{eqnarray*}
d_{b} & = & 2d_{a}\\
 & \Downarrow\\
\lim_{n\rightarrow\infty}c_{n} & = & \frac{\beta_{d_{b}}}{\alpha^2_{d_{a}}}
\end{eqnarray*}
In the second case, denoted as the super-exponential case, we have:
\begin{eqnarray*}
d_{b} & < & 2d_{a}\\
 & \Downarrow\\
\lim_{n\rightarrow\infty}c_{n} & = & 0
\end{eqnarray*}
And finally, in the third case, denoted as the sub-exponential or the polynomial case, we have:
\begin{eqnarray*}
d_{b} & > & 2d_{a}\\
 & \Downarrow\\
\lim_{n\rightarrow\infty}c_{n} & = & \mbox{sign}\left(\beta_{d_{b}}\right)\cdot\infty
\end{eqnarray*}
For all cases, from some point $c_{n}\approx\frac{\beta_{d_{b}}}{\alpha_{d_{a}}^{2}}n^{d_{b}-2d_{a}}$,
meaning that $\mbox{sign}\left(c_{n}\right)=\mbox{sign}\left(\beta_{d_{b}}\right)$.
Thus $\left|q_{n}\right|=\left|q_{n-1}+c_{n}q_{n-2}\right|$ is monotonically increasing.
Based on the observation that $\eta$ is a rational function of any tail $\tau_n$, it is enough to show that the above claims for the convergence rate apply for a tail $\tau_n$ for some $n$.

\FloatBarrier
\subsubsection{Exponential}
This section shows that the PCF converges exponentially in case that $d_{b} = 2d_{a}$ and under an additional demand on the leading coefficients of $a_n$ and $b_n$ is satisfied (creating a positive determinant as in Section \ref{sec:1Pred}).
The essence of the calculation below uses the fact that any PCF is a rational function of the value of its (infinite) tail. The rate of convergence of the PCF is derived from the rate of convergence of its tail.

We first analyze the tail of the PCF.
Since $c_{n} = \frac{\beta_{d_{b}}}{\alpha_{d_{a}}^{2}} + \mathcal{O}\left(\frac{1}{n}\right)$, we define $c=\frac{\beta_{d_{b}}}{\alpha_{d_{a}}^{2}}$, and get that from some point, $q_n$ can be estimated with:
\begin{eqnarray*}
\left(\begin{matrix}q_{n+1}\\
q_{n}
\end{matrix}\right) & = & \left(\begin{matrix}1 & c\\
1 & 0
\end{matrix}\right)^{n-k} \left(\begin{matrix}q_{k}\\
q_{k-1}
\end{matrix}\right)
\end{eqnarray*}
Therefore, $q_{n}$ approximates the $1$-periodic PCF case and the condition on the determinant from Section \ref{sec:1Pred} translates to:
\begin{eqnarray*}
4c+1 & > & 0\\
 & \Updownarrow\\
4\beta_{d_{b}} & > & -\alpha_{d_{a}}^{2}
\end{eqnarray*}

We can define a constant $\kappa_k$ that accounts for the beginning of the PCF, up to an index $k$, after which we assume $c_{i}\approx c$. Using this approximation, we get that the tail of the PCF converges to $\lambda_{+}$ (WLOG, we assume that
$\left|\lambda_{+}\right|>\left|\lambda_{-}\right|$), and therefore:
\begin{eqnarray*}
\eta_{n} & \approx & \kappa_k\lambda_{+}\frac{1-\left(\frac{\lambda_{-}}{\lambda_{+}}\right)^{n-k+1}}{1-\left(\frac{\lambda_{-}}{\lambda_{+}}\right)^{n-k}}
\end{eqnarray*}
We then receive that for large values of $n$:
\begin{eqnarray*}
\left|\eta-\eta_{n}\right| & \approx & \left|\kappa_k\lambda_{+}\right|\left|1-\frac{1-\left(\frac{\lambda_{-}}{\lambda_{+}}\right)^{n-k+1}}{1-\left(\frac{\lambda_{-}}{\lambda_{+}}\right)^{n-k}}\right|\\
 & = & \left|\kappa_k\lambda_{+}\right|\left|\frac{\lambda_{-}}{\lambda_{+}}\right|^{n-k} \underbrace{\left|\frac{1-\left(\frac{\lambda_{-}}{\lambda_{+}}\right)}{1-\left(\frac{\lambda_{-}}{\lambda_{+}}\right)^{n-k}}\right|}_{\left(\sum_{i=0}^\infty \left(\frac{\lambda_-}{\lambda_+}\right)^i\right)^{-1}<1}\\
 & \leq & \left|\kappa_k\right|\left|\lambda_{+}\right|\left|\frac{\lambda_{-}}{\lambda_{+}}\right|^{n-k}
\end{eqnarray*}
The resulting exponential decrease of the error at large $n$ can also be written as
\begin{eqnarray*}
\left|\eta-\eta_{n}\right| & \sim & \left|\frac{2c}{2c+1+\sqrt{1+4c}}\right|^{n-k}
\end{eqnarray*}
With a more careful calculation, the estimate of $\kappa_k$ can be improved to get a tighter bound on the error.

\FloatBarrier
\subsubsection{Super-exponential}
This section shows that when $d_{b} < 2d_{a}$, the PCF converges super-exponentially.
We assume WLOG that $\beta_{d_{b}}>0$. Then, $\exists k$ so that for all $n \geq k$:
\begin{eqnarray*}
c_{n} & \approx & \frac{\beta_{d_{b}}}{\alpha^{2}_{d_{a}}}n^{\overbrace{d_{b}-2d_{a}}^{<0}} 
\end{eqnarray*}
Also, since $\forall n>k \, : \,c_n>0$, $q_n>q_k$, which in turn results in:
\begin{eqnarray*}
\left|\eta-\eta_{n}\right| & \leq & \left|\frac{\prod_{i=1}^{n}c_{i}}{q_{n+1}q_{n}}\right| \leq \left|\frac{\prod_{i=1}^{n}c_{i}}{q_{k}^2}\right|  \\
& \approx & \kappa_{1} \left(\frac{\beta_{d_{b}}}{\alpha_{^{2}d_{a}}}\right)^{n-k} \left(\frac{n!}{k!}\right)^{d_{b}-2d_{a}}\\
 & = & \kappa_{2}\frac{\left(\frac{\beta_{d_{b}}}{\alpha_{^{2}d_{a}}}\right)^{n-k}}{\left(n!\right)^{2d_{a}-d_{b}}}
\end{eqnarray*}
Since $\frac{exp(n)}{n!}$ is decreasing super-exponentially, the desired result is obtained.

\subsubsection{Sub-Exponential}
The case satisfying the determinant constraint $4\beta_{d_{b}}  > -\alpha_{d_{a}}^{2}$ can be seen as a limit of the exponential convergence case with $c \to \infty$, therefore the derived convergence is sub-exponential. We believe this sub-exponential convergence to be polynomial. 

\FloatBarrier
\subsubsection{Improving the Convergence Rate by a Tail Estimation}

In the case of an exponentially converging PCF, we found that from
some point the tail is approximately:
\[
1+\frac{c}{1+\frac{c}{1+\ldots}}
\]
We calculated the convergence value of 1-periodic PCFs like this
earlier. Therefore, we can improve a PCF calculation by substituting
this tail at the final step. Empiric results display an improvement of a fixed number of digits
(for any large $n$). This in turn allowed us to improve the performance of the MITM-RF algorithm.

\FloatBarrier

\section{Further Information About the Descent\&Repel Method and Results }\label{app:descent_n_repel}

\begin{figure}[h]
\begin{center}
\includegraphics[width=\textwidth]{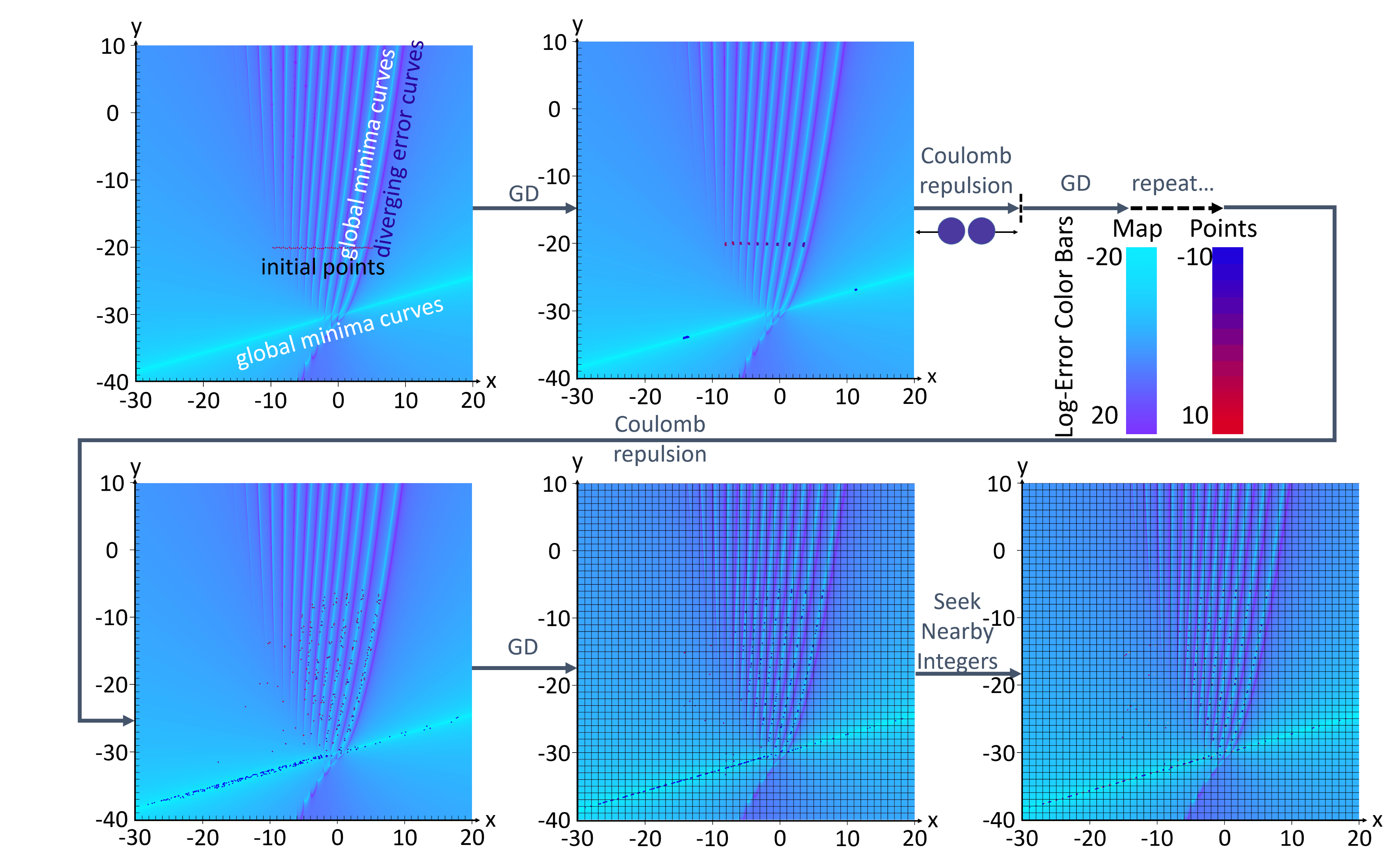}
\end{center}
\caption{\textbf{Descent\&Repel illustration}, as in Fig. \ref{fig:GDyahel}. Here the showcased scenario is that of the restoration of our previous result (found by the MITM-RF algorithm) $e=3+\frac{-1}{4+\frac{-2}{5+\frac{-3}{6+\ldots}}}$. The converging point is the one at $(x,y)=(4,-1)$.}
\label{fig:GD_map_conv}
\end{figure}

This section provides an additional example of the Descent\&Repel optimization process (in Fig. \ref{fig:GD_map_conv}), in addition to providing Table \ref{table:GDParams} with further information about the process presented in the main text (in Fig. \ref{fig:GDyahel}). The parameters chosen for Fig. \ref{fig:GDyahel} illustrate the optimization steps relatively clearly, however without converging to any real solution. In Fig. \ref{fig:GD_map_conv}, we present a similar illustration (Fig. \ref{fig:GD_map_conv}) presenting the convergence to $e=3+\frac{-1}{4+\frac{-2}{5+\frac{-3}{6+\ldots}}}$.

Below are the parameters required to reproduce the results in Fig. \ref{fig:GDyahel} and Fig. \ref{fig:GD_map_conv}.

\begin{table*}[h]
\begin{center}
\renewcommand{\arraystretch}{1.0}
\begin{tabular}{|l|l|l|}
\specialrule{.1em}{.05em}{.05em} 
\textbf{Fig. Number} & \textbf{Parameters}            & \textbf{Values} \\
\specialrule{.1em}{.05em}{.05em} 
    Fig. \ref{fig:GDyahel} & $a(n)$ & $n$ \\ \cline{2-3}
        & $b(n)$ & $n^2+ny+x$ \\ \cline{2-3}
        & $x$ range & $[-20,20]$ \\ \cline{2-3}
        & $y$ range & $[-20,20]$ \\ \cline{2-3}
        & fraction depth & 10 \\ \cline{2-3}
        & constant & $\pi$ \\ \cline{2-3}
        & initial points & 600, uniform at $y\in[-15, 15]$ and $x=10$ \\ \cline{1-3}
    Fig. \ref{fig:GD_map_conv} & $a(n)$ & $n+x$ \\ \cline{2-3}
        & $b(n)$ & $-n+y$ \\ \cline{2-3}
        & $x$ range & $[-30,20]$ \\ \cline{2-3}
        & $y$ range & $[-40,10]$ \\ \cline{2-3}
        & fraction depth & 20 \\ \cline{2-3}
        & constant & $e-3$ \\ \cline{2-3}
        & initial points & 500, uniform at $y = -20$ and $x\in[-10,5]$ \\ \cline{1-3}
\specialrule{.1em}{.05em}{.05em} 
\end{tabular}
\end{center}
\caption{\textbf{Execution settings required to reproduce the Descent\&Repel maps} of Fig. \ref{fig:GDyahel} and Fig. \ref{fig:GD_map_conv}). Here, $a,b$ are similar to the polynomials $\alpha,\beta$ that define the PCF, but the RF is of the form $\frac{b_0}{a_0+\frac{b_1}{a_1+\ldots}}$.}
\label{table:GDParams}
\end{table*}

\FloatBarrier
\newpage

\section{Example Proofs of Ramanujan Machine Results}\label{app:ConjProofs}

This section summarizes several proofs for the PCFs found by the Ramanujan Machine project.
We periodically update the paper and the website with more proofs suggested by the wider community after we verify them. Each case requires different identities, and it remains to be seen whether all the cases discovered so far by the algorithms of the Ramanujan Machine can be proven by existing math, or eventually require inventing new mathematical techniques.

\subsection{Proof for \texorpdfstring{$e = 3 + \frac{-1}{4+\frac{-2}{5+\frac{-3}{6+\ldots}}}$}{eq-proof-e}}

The proof for the aforementioned continued fraction was presented in \cite{lu2019elementary} (can also be proven with a variant of the Euler continued fraction). The proof relies on the following feature of PCFs \cite{jones1984continued}. Define the auxiliary series on the $a_{n}$ and $b_{n}$ polynomials:


\begin{equation*}
  A_{n} = 
  \begin{cases}
    b_{n}A_{n-1}+a_{n}A_{n-2} & \text{if $n>1$} \\
    b_{0} & \text{if $n=0$} \\
    1 & \text{if $n=-1$}
  \end{cases}
  \quad\quad,\quad\quad
    B_{n} = 
  \begin{cases}
    b_{n}B_{n-1}+a_{n}B_{n-2} & \text{if $n>1$} \\
    1 & \text{if $n=0$} \\
    0 & \text{if $n=-1$}
  \end{cases}
\end{equation*}
The value of the PCF is the limit of the ratio between the two auxiliary sequences $PCF(\alpha,\beta) = \lim_{n \to \infty} \frac{A_{n}}{B_{n}}$. In the case of this PCF, the auxiliary series are: 

\begin{equation*}
  A_{n} = 
  \begin{cases}
    (n+3)A_{n-1}-nA_{n-2} & \text{if $n>1$} \\
    3 & \text{if $n=0$} \\
    1 & \text{if $n=-1$}
  \end{cases}
  \quad\quad,\quad\quad
    B_{n} = 
  \begin{cases}
    (n+3)B_{n-1}+-nB_{n-2} & \text{if $n>1$} \\
    1 & \text{if $n=0$} \\
    0 & \text{if $n=-1$}
  \end{cases}
\end{equation*}
By using induction and observing that $A_{n}$ is a shifted version of sequence A001339 in \cite{sloane2003line} the author of \cite{lu2019elementary} gets that:

\begin{equation*}
    B_{n} = \frac{((n+1)!)^{2}}{n!} = (n+1)\cdot(n+1)!
      \quad\quad,\quad\quad
    A_{n} = \sum_{k=0}^{n+1}(k+1)!\binom{n+1}{k}
\end{equation*}
From there, the authors calculate the limit of the ratio and conclude the proof:
\begin{equation*}
    \lim_{n \to \infty} \frac{A_{n}}{B_{n}} = \lim_{n \to \infty} \frac{n+2}{n+1}\sum_{k=0}^{n+1}\frac{1}{k!} - \frac{1}{n+1}\sum_{k=0}^{n}\frac{1}{k!} = e
\end{equation*}

We expect there to be a more general proof that covers any PCF in which both polynomials are of degree one, but it seems to require certain identities of the Lerch Transcendent (see page 475 in \cite{perron1913lehre}) and requires further research.

\subsection{Proof for two conjectures regarding \texorpdfstring{$\pi$}{pi}}
Both proofs use identities for the ratio of generalized hypergeometric functions, which can be written in two different PCF forms:
\setcounter{equation}{0}
\begin{equation}
    \frac{c\cdot_{2}F_{1}(a,b;c;x)}{_{2}F_{1}(a+1,b;c+1;x)} = c+\frac{(a-c)bx}{(c+1)+\frac{(b-c-1)(a+1)x}{(c+2)+\frac{(a-c-1)(b+1)x}{(c+3)+\frac{(b-c-2)(a+2)x}{(c+4)\dots}}}}
\label{eq:gauss1}
\end{equation}
\begin{equation}
\frac{c\cdot_{2}F_{1}(a,b;c;x)}{_{2}F_{1}(a+1,b;c+1;x)} = c + (1+a-b)x-\frac{(a+1)(1+c-b)x}{(c+1)+(2+a-b)x-\frac{(a+2)(2+c-b)x}{(c+2)+(3+a-b)x-...}}
\label{eq:gauss2}
\end{equation}

Similar methods (Using Gauss' continued fraction), can also be used to prove all the results we found so far for variations on $\pi$ [see \ref{table:ResTable_pi}].However, we do not know any similar technique that proves the results for $\pi^2$, $\zeta(3)$, or Catalan's constant.

\subsubsection{\label{sub:proof-of-general-pi} Proof for \texorpdfstring{$ \frac{2^{2\cdot z +1}}{\pi\binom{2\cdot z}{z}} = 1+\frac{1\cdot(2\cdot z-1)}{4+\frac{2\cdot(2\cdot z-3)}{7+\frac{3\cdot(2\cdot z-5)}{..}}} $}{proof-pi-2}}
By using Eq. \ref{eq:gauss2}, we multiply both sides by 2 and substitute:
$a = -\frac{1+2z}{2}, b=c=\frac{1+2z}{2}, x=\frac{1}{2}$
To get:
$$ 1+\frac{2z-1}{4+\frac{2\cdot(2z-3)}{7-...}} = \frac{2b\cdot_{2}F_{1}(-b,b;b;\frac{1}{2})}{_{2}F_{1}(1-b,b;1+b;\frac{1}{2})}$$
Using the following identities for $_2F_1$:
$$_2F_1\left(\alpha,\beta;\beta;z\right) = (1-z)^{-\alpha}$$
and:
$$ _2F_1\left(\alpha, 1-\alpha, \gamma, \frac{1}{2}\right)= \frac{\Gamma\left(\frac{1}{2}\gamma\right) \Gamma\left(\frac{1}{2}(1+\gamma)\right)}{\Gamma\left(\frac{1}{2}(\gamma+ \alpha)\right) \Gamma\left(\frac{1}{2}(1+\gamma-\alpha)\right)}  $$
We then get:
$$  1+\frac{2z-1}{4+\frac{2\cdot(2z-3)}{7-...}} = 2^b\cdot 2b \cdot \frac{\Gamma\left(b+\frac{1}{2}\right)}{\Gamma\left(\frac{b+1}{2}\right)\Gamma\left(\frac{b+2}{2}\right)} =  \frac{2\Gamma\left(b+\frac{1}{2}\right)}{\sqrt{\pi}\Gamma\left(b\right)} = \frac{2\Gamma\left(z+1\right)}{\sqrt{\pi}\Gamma\left(z+\frac{1}{2}\right)} = \frac{2^{2\cdot z +1}}{\pi\binom{2\cdot z}{z}} $$

\subsubsection{Proof for \texorpdfstring{$\frac{4}{\pi-2} = 3 + \frac{1\cdot3}{5+\frac{2\cdot4}{7+\frac{3\cdot5}{9+\ldots}}}$}{proof-pi-1}}
By using the identity of Eq. \ref{eq:gauss1} and substituting $a=0,b=\frac{1}{2},c=\frac{3}{2},x=-1$ we arrive at:
\begin{equation*}
    3+\frac{1\cdot3}{5+\frac{2\cdot4}{7+\frac{3\cdot5}{9+\frac{4\cdot6}{11+\dots}}}} = 3\frac{1}{_{2}F_{1}(1,\frac{1}{2};\frac{5}{2};-1)}  = \frac{4}{\pi - 2}
\end{equation*}

\subsection{\label{sub:DaughertyProofs}Proof of \texorpdfstring{$\frac{3}{3 - e} = 11 - \frac{10}{29 - \frac{28}{55 - \frac{54}{89 - \frac{88}{..}}}}$}{dougherty1} and \texorpdfstring{$1+\frac{e}{e - 2} = 5 - \frac{4}{19 - \frac{18}{41 - \frac{40}{71 - \frac{70}{..}}}}$}{dougherty2}}

These PCFs were proven by Dougherty and Zeilberger using the following identity that they proved in \cite{ZeilbergerDougherty-Bliss2020}. Following their notation, we shall denote:
$$[an^2+bn+1 : -an^2 -bn] = \frac{F(a,b) + 2(2a+b)(a+b+1)}{F(a,b) + 2(2a+b)}$$
where:
$$ [a(n) : b(n)] = a(1) + \frac{b(1)}{a(2)+\frac{b(2)}{a(3)+\frac{b(3)}{...}}}$$
$$ F(a,b) = 2\sum_{k\geq0}{\frac{1}{(k+2)!(3+b/a)^{\overline{k}}a^k}} $$
$$ x^{\overline{k}} \text{ is the rising factorial also known as } x_{\left(k\right)} \text{ i.e. } x(x+1)...(x+k-1)$$

\subsubsection{\texorpdfstring{$\frac{3}{3 - e} = 11 - \frac{10}{29 - \frac{28}{55 - \frac{54}{89 - \frac{88}{..}}}}$}{dougherty1}}
In this case:
 $$[a(n) : b(n)] \text{ where}:\quad a(n) = 4n^2+6n+1 \quad, \quad b(n)=-4n^2-6n $$
hence for the mentioned identity we use $a=4,\quad b=6$.
 $$ F(4,6) = 2\sum_{k\geq0}{\frac{1}{(k+2)!(3+3/2)^{\overline{k}}4^k}} = \frac{28 (30 - 11 e)}{e} $$
(Using the steps similar to the ones used by Dougherty and Zeilberger).\newline
When we plug this result into the rest of the expression we receive:
$$\frac{\frac{28(30 - 11e)}{e} + 2(2\cdot 4+6)(4+6+1)} {\frac{28(30 - 11e)}{e} + 2(2\cdot4+6)} = \frac{3}{3-e}  $$

\subsubsection{\texorpdfstring{$1+\frac{e}{e - 2} = 5 - \frac{4}{19 - \frac{18}{41 - \frac{40}{71 - \frac{70}{..}}}}$}{dougherty2}}

We "roll up" the continued fraction by 1 step, using the transformation $f(x)=\frac{2}{x}-1$ to both sides of the equation, and get:
$$\frac{1}{1 - e} = -1+\frac{2}{5 - \frac{4}{19 - \frac{18}{41 - \frac{40}{71 - \frac{70}{..}}}}}$$
which can be written as
 $$ [a(n) : b(n)] \text{ where}:\quad a(n) = 4n^2-6n+1 \quad, \quad b(n)=-4n^2+6n $$
Therefore, for the mentioned identity we use $a=4,\quad b=-6$ and write
 $$ F(4,-6) = 2\sum_{k\geq0}{\frac{1}{(k+2)!(3-3/2)^{\overline{k}}4^k}} = \frac{4(e - 2)}{e} $$
When we plug this result into the rest of the expression we receive:
$$\frac{\frac{4(e - 2)}{e} + 2(2\cdot 4+6)(4-6+1)} {\frac{4(e - 2)}{e} + 2(2\cdot4-6)} = \frac{1}{1 - e}  $$